# Observational and Experimental Insights into Machine Learning-Based Defect Classification in Wafers

Kamal Taha, *Senior Member*
Department of Computer Science, Khalifa University, Abu Dhabi, UAE. E-mails: kamal.taha@ku.ac.ae

***Abstract*— This survey paper offers a comprehensive review of methodologies utilizing machine learning (ML) classification techniques for identifying wafer defects in semiconductor manufacturing. Despite the growing body of research demonstrating the effectiveness of ML in wafer defect identification, there is a noticeable absence of comprehensive reviews on this subject. This survey attempts to fill this void by amalgamating available literature and providing an in-depth analysis of the advantages, limitations, and potential applications of various ML classification algorithms in the realm of wafer defect detection. An innovative taxonomy of methodologies that we present provides a detailed classification of algorithms into more refined categories and techniques. This taxonomy follows a three-tier structure, starting from broad methodology categories and ending with specific techniques. It aids researchers in comprehending the complex relationships between different algorithms and their techniques. We employ a rigorous Observational and experimental evaluation to rank these varying techniques. For the Observational evaluation, we assess techniques based on a set of four criteria. The experimental evaluation ranks the algorithms employing the same techniques, sub-categories, and categories. Also the paper illuminates the future prospects of ML classification techniques for wafer defect identification, underscoring potential advancements and opportunities for further research in this field**.

***Note to Practitioners* — The scattered information on ML methods for wafer defect detection hinders a complete understanding of optimal techniques and their varying effectiveness. This survey paper attempts to rectify this issue by providing an in-depth review of ML approaches used for identifying and classifying defects on wafers. Our objective is to amalgamate available literature to underscore the advantages, drawbacks, and potential uses of various ML algorithms**.

## I. INTRODUCTION

Integrated circuits (ICs) are densely packed electronic circuits on silicon chips, produced from wafers. They are crucial for advanced technologies like AI [1], IoT [2], the automotive industry, and 5G networks [3]. This significance has led to a growing demand for semiconductors, necessitating efficient manufacturing automation by fabrication companies.

The production of high-quality semiconductors involves reducing defects during the wafer fabrication process, as these defects can lead to chip failure [4]. Effective defect monitoring is vital for production yield in chip fabrication, with traditional manual inspections being costly and less accurate. Image processing and machine learning techniques are emerging as more cost-effective and accurate solutions.

Wafer Bin Maps (WBM) play a crucial role in this context. They visually represent and categorize defective dies on a wafer, using colors to indicate their status [5]. These maps are essential for diagnosing defects, spotting patterns, uncovering causes, and tracking semiconductor production.

Defective chips tend to cluster and show spatial correlations that can indicate the causes of flaws [6]. By studying these patterns, improvements in process engineering can be achieved, thereby enhancing product quality and increasing the yield of defect-free chips [7, 8, 9]. Effective defect monitoring is key to production yield in chip fabrication, with traditional manual inspections proving costly and less accurate [10]. Image processing and machine learning techniques offer more cost-effective and accurate solutions [11, 12].

Machine learning (ML) algorithms, known for their ability to process and learn from vast datasets, have found widespread application across numerous sectors, prominently including the field of wafer defect detection. These algorithms harness substantial computational power, allowing them to efficiently analyze intricate and often subtle patterns within the data. This capability is particularly crucial in wafer defect detection, where the identification of minute and complex defect patterns is essential for quality control and assurance in semiconductor manufacturing. By utilizing advanced ML techniques, these systems can discern defects that might be imperceptible to human inspectors, thereby significantly enhancing the accuracy and reliability of the defect identification process.

The adoption of deep learning, a subset of ML, is widespread due to technological advancements, significantly benefiting the semiconductor industry by improving flaw detection and analysis [13, 14, 15, 16, 17]. It is becoming key in wafer defect identification due to its superiority over traditional methods. These include better recognition of complex patterns in wafer defects, adaptability in learning from data, and handling large datasets more effectively [18]. Deep learning reduces the need for manual feature extraction and expert intervention, leading to more autonomous, efficient, and error-reduced operations. Its accuracy and sensitivity are higher. Moreover, deep learning models offer scalability and versatility, easily updating for new defect types, aligning with the evolving complexity of semiconductor manufacturing and its shift towards advanced, automated processes.

Although ML has shown significant efficiency in identifying defects in wafers, there is a notable lack of thorough reviews in this field. Our work aims to bridge this by providing a thorough survey of ML classification algorithms, detailing their techniques, sub-categories, and categories. This taxonomy facilitates a clearer assessment and comparison of algorithms, highlighting their pros and cons, and sets a foundation for future research to refine and evaluate new ML approaches.

This survey not only presents a detailed framework for categorizing ML classification algorithms but also includes ***Observational*** and ***experimental*** evaluations to measure the effectiveness of different approaches. Our ***Observational evaluation*** focuses on techniques for identifying wafer defects based on four criteria. Through ***experimental evaluation***, we compare and rank various algorithmic categories and techniques, including those utilizing the same technique, different techniques within the same sub-category, and different sub-categories.

## A. Key Contributions

### 1. Providing Methodology-Based Taxonomy

We introduce a methodology-based taxonomy that categorizes defect classification methods into three principal categories: type-based, label-based, and agent-based (see Fig. 1).

- **Type-Based Methods**: These methods are further divided into single-type and multi-type methods. Single-type methods focus on identifying one specific kind of defect, whereas multi-type methods are capable of identifying multiple defect types concurrently.
- **Label-Based Methods**: Classified into single-label and multi-label methods, this category addresses the output granularity of the classification process. Single-label methods assign one label per instance, ideal for scenarios where each wafer can have only one type of defect. In contrast, multi-label methods allow for the assignment of multiple labels to a single instance, accommodating the complexity of real-world scenarios where multiple defects may coexist on a single wafer.
- **Agent-Based Methods**: These methods are categorized into single-agent and multi-agent systems. Single-agent methods employ a solitary model or algorithm to perform the classification, whereas multi-agent methods use a collaborative approach among multiple agents or algorithms. This division underscores the potential for complex problem-solving strategies and enhanced performance through collaboration.

### 2. Providing Observational Evaluations

We perform Observational evaluations to gauge the efficacy of different methodologies. Specifically, we evaluate the methods in terms of Complexity, Performance, Robustness, and Limitations**.**

### 3. Providing Experimental Evaluations

Through detailed experimental evaluations, our manuscript compares and ranks various algorithmic categories and techniques. This includes comparisons of algorithms that utilize the same single-label, multi-label, single-type, multi-label, single-agent, and multi-agent techniques.

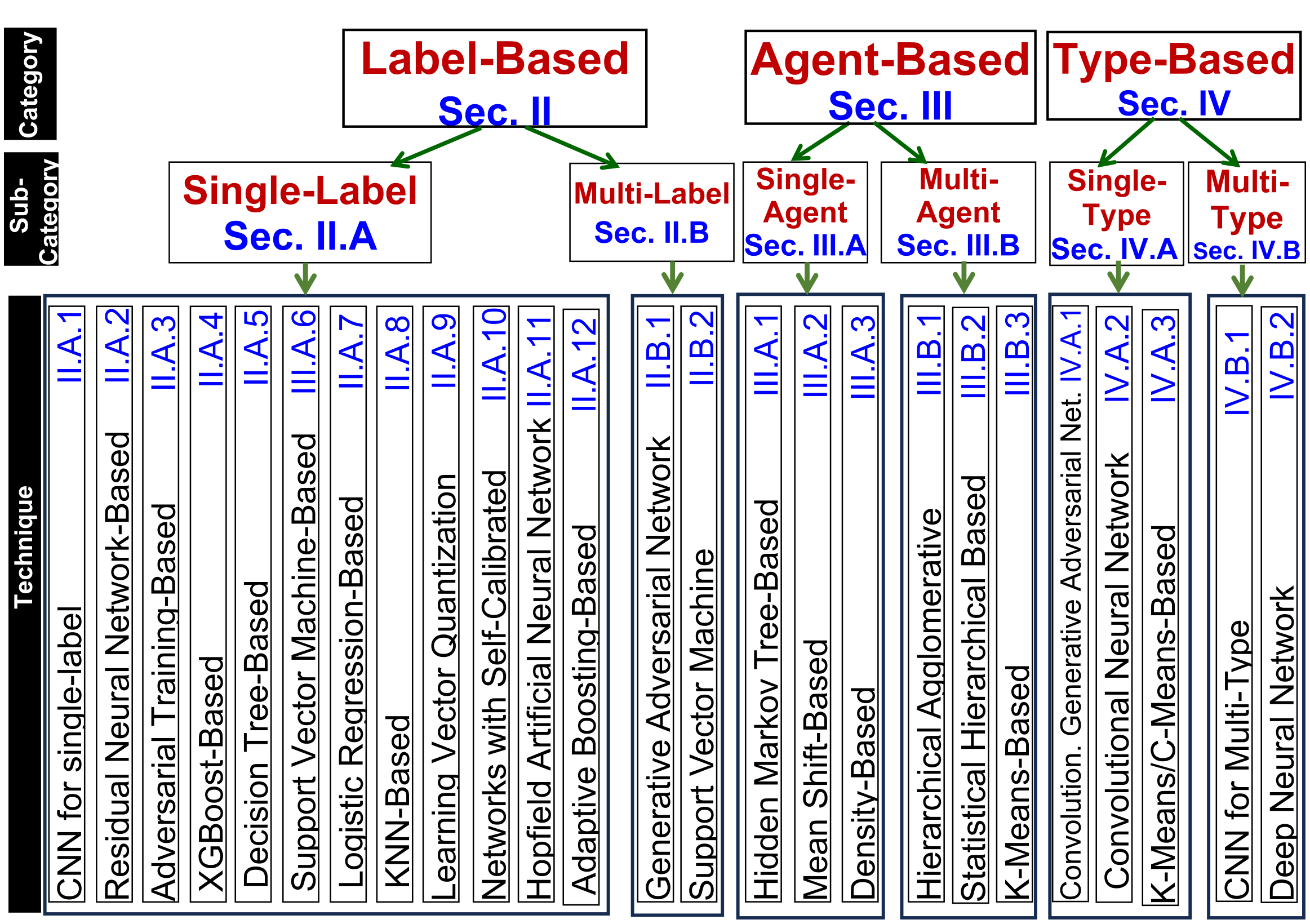

**Fig. 1:** The figure illustrates our hierarchical methodology-based taxonomy for classifying ML classification algorithms utilized in wafer defect identification. The taxonomy categorizes the algorithms into fine-grained classes, progressing from methodology category to methodology sub-category, and finally methodology technique. The figure provides the corresponding section numbers in the manuscript.

## II. Label-Based Classification Category

### A. Single-Label Classification Sub-Category

#### 1. Convolutional Neural Network (CNN) for Single-Label Wafer Defect Technique

This technique applies convolutional layers to extract and learn features from wafer images in a hierarchical manner. This structure allows the CNN to efficiently discern intricate patterns and characteristics specific to different types of wafer defects, leading to accurate classification based on the learned features.

*a) The components of the Technique*

(1) Layered Architecture in CNNs: In CNNs designed for wafer defect classification, the layered architecture is pivotal. By leveraging a combination of convolutional and pooling layers, the network adeptly identifies intricate features indicative of defects, such as scratches, irregular patterns, and anomalies on the wafer surface. These layers work in concert to extract and highlight features critical for ensuring the quality and reliability of wafers, (2) Convolutional Layers: At the forefront of wafer defect detection, convolutional layers utilize specialized filters to scrutinize wafer images for essential features like edges and textures. This early-stage feature extraction is vital, as it allows the CNN to pinpoint potential defect areas by recognizing patterns that deviate from the norm, thereby facilitating early intervention in the manufacturing process, (3) Pooling Layers for Data Reduction: Given the vast datasets typical in semiconductor manufacturing, pooling layers in CNNs serve the critical function of data reduction. By downsampling the feature maps, these layers significantly cut down the computational load and memory requirements, (4) Fully Connected Layers: CNNs use fully connected layers after convolutional and pooling layers to classify wafer detected features into specific defect types, and (5) Softmax Layer for Probability Distribution: The softmax layer in CNNs for wafer defect classification provides probabilities for each defect type. Fig. 2 depicts the CNN procedure for detecting wafer defects.

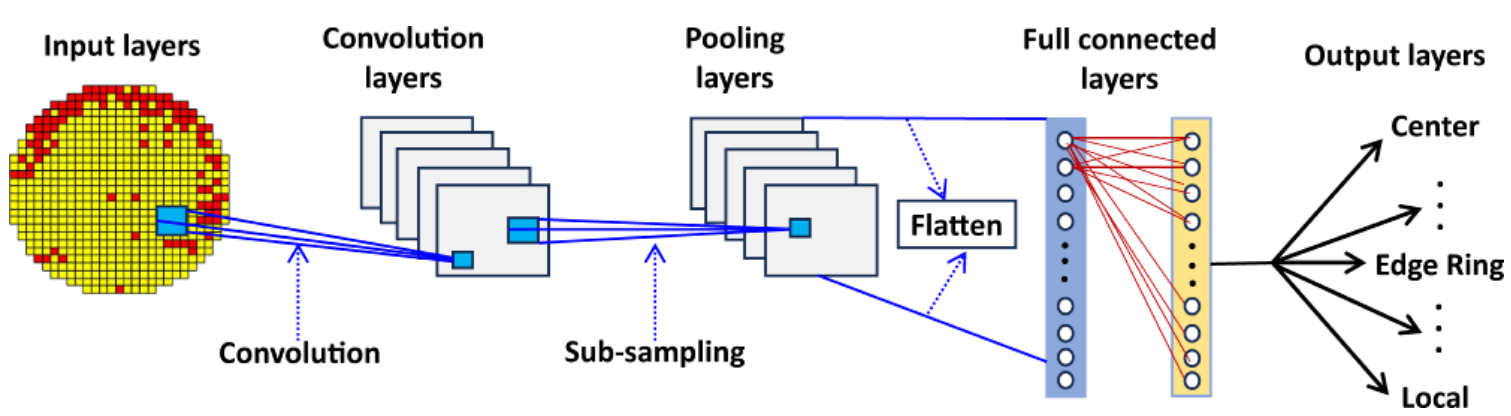

**Fig. 2:** Wafer defect pattern detection model diagram using CNN.

*b) The rationale behind the usage of the technique*

(1) Local Pattern Recognition: The technique excels in detecting specific patterns and spatial relationships in small image areas, crucial for pinpointing wafer defects, (2) Automatic Feature Learning: CNNs can autonomously identify wafer image features, including basic elements like edges and textures, and complex shapes, aiding in distinguishing wafer defects, (3) Hierarchical Modeling: With multiple layers, CNNs learn features at varying complexity levels, enabling them to capture intricate patterns indicative of wafer defects, and (4) Pretraining for Generalization: Training on large datasets allows CNNs to understand diverse wafer image features, making them adaptable for tasks like defect detection, even with limited labeled data.

*c) The conditions for the technique's optimal performance*

(1) Use Pre-trained Models: Employ pre-trained CNN models, such as those trained on ImageNet, and fine-tune them for specific datasets like wafer defect data, (2) Choose the Right CNN Architecture: Select a CNN architecture like ResNet that can capture data intricacies while being computationally efficient, (3) Control CNN Depth: Balance the CNN depth to capture complex patterns without overfitting the data, and (4) Optimize Hyperparameters: Choose optimizer algorithms (e.g., Adam) and fine-tune hyperparameters like learning rates, batch sizes, and regularization methods (e.g., dropout).

*d) Research Papers that have Employed the technique*

Chen et al. [19] introduced a predefined CNN model along with transfer learning, which leverages pre-trained parameters to assist the network in capturing the fundamental patterns found in the wafer map defect pattern. Shen and Zheng [20] introduced a deep transfer learning model called JFLAN that uses CNNs to extract transferable features of wafer maps. It offers a unique feature learning approach using transfer learning. It employs multilayer domain adaptation through adversarial training. A. R and James [21] developed an automated system for wafer defect classification using a CNN combined with a memristor crossbar structure. Pre-trained neural network weights are implemented within the crossbar structure, and classification is based on softmax layer.

*e) Case Studies and Application of the Technique*

Engineers at GlobalFoundries [22] conducted a study comparing a traditional Support Vector Machine (SVM) approach, commonly used in computer vision, with a 4-layer deep CNN for wafer test map classification. Their aim was to enhance accuracy in identifying low yield defects. The SVM model necessitated feature engineering for training, whereas the CNN leveraged existing image datasets, undergoing 120 epochs. Both models were trained using 300 to 500 manually labeled images for each of the 12 unique wafer map signatures. CNN markedly outperformed the SVM in accuracy, averaging 90% across the signatures with a strong sensitivity to pattern shape.

### 2. Residual Neural Network (ResNet) Technique

The technique identifies and classifies defects in wafers through a hierarchical feature extraction process. It uses residual blocks, which allow for the training of very deep networks by addressing the vanishing gradient problem through skip connections that enable the flow of gradients directly through the network layers.

*a) The Major Components of the Technique*

(1) Input Layer: It accepts images of wafers, which are typically represented as multi-dimensional arrays, (2) Batch Normalization Layer: It normalizes the output of the input layer by subtracting the batch mean and dividing by the batch standard deviation, (3) Convolutional Layer: It performs feature extraction by applying a convolution operation between the input data and a set of learnable filters, capturing spatial hierarchies in wafer images. It can detect patterns, (4) Residual Modules: They allow the network to learn identity functions, ensuring that deeper network layers can perform at least as well as shallower ones. They contain two or three convolutional layers, each followed by a batch normalization layer, with a skip connection that adds the input of the block to its output, to mitigate vanishing gradient, (5) Fully Connected Layer: Makes predictions based on the features extracted through the network. It transforms the learned features into final outputs, such as class scores for classification tasks. Fig. 3 depicts this procedure

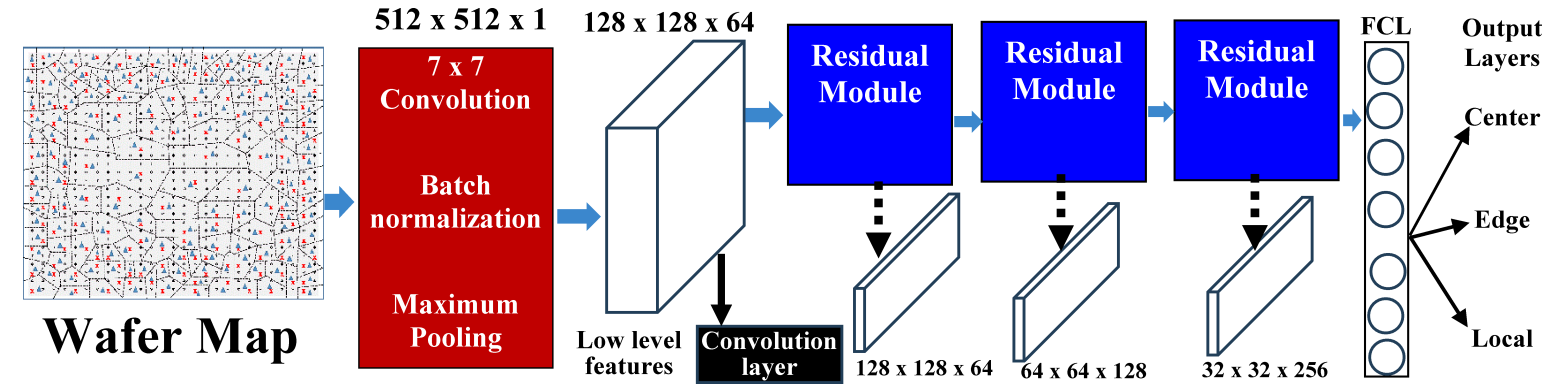

**Fig. 3:** Wafer defect pattern detection model diagram using ResNet.

*b) The rationale behind the usage of the technique*

(1) Complexity of Defects: Wafers can exhibit a wide range of defect types. These defects vary greatly in terms of size, shape, and appearance. ResNet's architecture, characterized by deep layers and skip connections, is adept at learning from such complex and variable data. It can extract hierarchical features that capture both the subtle nuances and the distinct differences among various defect types, (2) Deep Feature Representation: Many wafer defects are subtle and can be easily missed or misclassified by less sophisticated models. ResNet's deep structure allows for the learning of intricate patterns and features that are crucial for identifying these less obvious defects. The residual connections help in mitigating the vanishing gradient problem.

*c) The conditions for the technique's optimal performance*

(1) Model Depth: Choose an appropriate ResNet variant (e.g., ResNet-50, ResNet-101) based on the complexity of the wafer defect types. More layers can capture complex wafer defect patterns but require more data to avoid overfitting, (2) Input Resolution: Configure the network to accept the native resolution of the preprocessed wafer images, ensuring that small defect features are not lost in down-sampling, and (c) Feature Extraction Focus: Modify layers to focus on feature extraction capabilities for small and subtle wafer defect features.

*d) Research Papers that have Employed the technique*

Li and Wang [23] introduced an enhanced mask R-CNN model that combines the residual network and feature pyramid network to enhance the recognition capability of small targets. Amogne et al. [24] introduced the Opt-ResDCNN model, a deep convolutional neural network with residual blocks. This model was designed for classifying defect patterns in wafer maps. Inspired by ResNet, the method enhances the model by incorporating additional convolutional layers and residual blocks.

*e) Case Studies and Application of the Technique*

Intel [25] has harnessed this technology in a system implemented across its assembly and test factories, leveraging computer vision and ResNet50. This system conducts inline inspections to identify defects, outperforming older offline methods in efficiency and effectiveness. Central to ensuring the quality and dependability of Intel's new product lines, this technology has been integrated throughout Intel's manufacturing facilities. This system has successfully identified a range of defects in the wafer-thinning process, including indentations, scratches, stains, cracks, bubbles, and discrepancies in wafer and mount alignment.

## 3. Adversarial Training Technique

The technique involves training a model using adversarially generated examples, specifically tailored to mimic wafer defects, to enhance the robustness and accuracy of defect classification. This process involves iteratively modifying wafer images to introduce or exaggerate features that resemble manufacturing defects, thereby challenging the model to learn distinguishing characteristics effectively. This adversarial approach helps in significantly improving the model's ability to classify wafer defects accurately by reinforcing its defenses against subtle, yet critical, variations in defect appearances.

*a) The Components of the Technique*

(1) Adversarial Example Generation: This involves creating perturbed versions of the original wafer images in the dataset. These perturbations are usually small, often imperceptible to the human eye, but are designed to mislead the classification model. Techniques like the Fast Gradient Sign Method, Projected Gradient Descent, and Carlini & Wagner attacks can be used to generate these adversarial examples. The idea is to simulate potential attacks or scenarios where the model might fail, focusing specifically on types of alterations that could mimic or obscure wafer defects, and (2) Model Architecture: For wafer defect classification, CNNs architecture should be designed or chosen to balance accuracy in wafer defect detection with resilience to adversarial examples.

*b) The rationale behind the usage of the technique*

(1) Complexity of Wafer Defect Patterns: Adversarial training introduces difficult, synthetic examples of wafer defect patterns during the training phase, which helps the model to learn a more comprehensive representation of defect patterns, (2) Generalization Across Variations: Wafer manufacturing processes and tools evolve over time, leading to new types of defects or variations in existing ones. Adversarial training can simulate these variations, ensuring the model remains effective even as the manufacturing environment changes, and (3) Handling Subtle Defects: Some wafer defects are subtle and can be easily missed or misclassified by models trained on less challenging datasets. By generating adversarial examples that mimic these subtleties, the model can learn to identify wafer defects.

*c) The conditions for the technique's optimal performance*

(1) Example Generation: Realistic adversarial examples should be crafted to mimic potential variations and anomalies that can occur in wafer production environments. This includes simulating wafer defects of varying sizes, shapes, and types that might not yet have been encountered in real production settings but could theoretically exist, ensuring the model can generalize well to new, unseen defects, and (2) Balance Between Robustness and Accuracy: This balance ensures that the system can accurately identify true wafer defects without being misled by noise or minor surface irregularities that do not impact the wafer's functionality.

*d) Research Papers that have Employed the technique*

Yu et al. [26] introduced DTWAN, an adaptive transfer learning framework leveraging adversarial training. It utilizes multi-stage optimization, incorporating maximum mean discrepancy (MMD), cross entropy, and adversarial loss. A generative adversarial algorithm is crafted within this framework to assist the model in extracting universal features from both source and target domains. DTWAN efficiently transfers essential knowledge from the source to the target domain, reducing data collection costs and enhancing the industrial utility of the recognition model. Wang et al. [27] introduced an adaptive balancing generative adversarial training technique for imbalanced learning by combining adversarial training and domain adaptation. Liu et al. [28] introduced a method using generative adversarial training for simulating defective samples, addressing the scarcity of such samples in production. They designed a network with an encoder-decoder structure, training it alongside a discriminative network under a novel regional training strategy that focuses on defective areas first. The approach enhances defect-free regions via wavelet fusion, efficiently generating defects of specific shapes and types with minimal training samples, while also providing precise pixel-wise ground truth.

## 4. XGBoost-Based Technique

The technique leverages the gradient boosting framework to efficiently identify and classify various types of defects on wafers. By employing an ensemble of decision trees, XGBoost analyzes the features extracted from wafer images to learn and predict the specific type of defect present. It utilizes advanced regularization techniques to prevent overfitting.

*a) The Major Components of the Technique*

(1) Objective Function: Utilizes "multi-class logloss" to cater to the multi-class nature of wafer defect types, optimizing the model to reduce misclassifications across various defect categories, (2) Gradient and Hessian Calculations: Captures the complex patterns indicative of different wafer defects, ensuring precise updates to the model with each iteration, (3) Tree Ensemble: Builds a series of decision trees sequentially, each correcting the previous trees' errors, capturing the multifaceted nature of defect signatures on wafers, and (4) Feature Importance: Identifies key features that influence defect classification (e.g., defect size, location), providing insights into defect characteristics.

*b) The rationale behind the usage of the technique*

(1) Handling Imbalanced Data: Wafer defect datasets often exhibit a class imbalance. XGBoost can handle this imbalance efficiently through its scale_pos_weight parameter, which helps in tuning the algorithm to improve performance on the minority class without losing accuracy on the majority class, (2) Feature Importance: XGBoost provides built-in support for assessing feature importance, which can be crucial for wafer defect classification, (3) Flexibility: XGBoost supports various objective functions and evaluation criteria, allowing for customization tailored to the specific characteristics of wafer defect classification. For example, it can be adjusted to focus on precision or recall, depending on the cost of different types of classification errors in the wafer.

*c) The conditions for the technique's optimal performance*

(1) Feedback Loop: Implement a feedback system where model predictions are periodically reviewed by experts, and the model is retrained with updated labels to adapt to new types of defects or changes in the manufacturing process, and (2) Model Interpretability: Use SHAP (SHapley Additive exPlanations) values or other interpretability tools to understand how different features impact the model's predictions.

*d) Research Papers that have Employed the technique*

Yuan-Fu [29] utilized XGBoost and CNN to tackle wafer map retrieval tasks and the classification of defect patterns. Xu et al. [30] introduced an enhanced multi-batch wafer yield prediction model based on XGBoost aimed at boosting production efficiency and minimizing wafer defects. They developed a multi-task learning approach for batch feature extraction and established a fusion training mechanism to facilitate predictive output.

*e) Case Studies and Application of the Technique*

Intel's integration of XGBoost on their CPUs for wafer defect detection is a pivotal enhancement in semiconductor manufacturing [31]. Leveraging XGBoost's strengths, particularly its adeptness at handling large datasets and accelerating data processing phases, this approach aligns perfectly with the demands of wafer production's high-precision environment. The synergy of XGBoost with Intel's processing power significantly boosts efficiency, enabling faster and more accurate analysis of extensive data. This advancement is crucial in the rapid-paced, accuracy-focused world of semiconductor manufacturing, leading to markedly improved wafer defect detection.

### 5. Decision Tree-Based Technique

The technique employs a hierarchical, tree-like model to categorize different types of wafer defects by making decisions based on the attributes of wafer images, such as texture, shape, and defect features. At each node of the tree, the algorithm chooses the feature that best separates the data into classes, creating branches until it reaches a decision or leaf node that represents a defect type. It simplifies the complex decision-making process by breaking down the classification problem into simpler decisions.

*a) The Major Components of the Technique*

(1) Decision Nodes: Points in the tree where the data is split according to certain criteria related to wafer defects, like defect types and their characteristics (e.g., diameter), (2) Leaf Nodes: The end points of the decision tree that provide the final classification of the wafer as either defective or non-defective based on the criteria defined in the decision nodes, specifying the type and severity of defects if present, (3) Splitting Criteria: The rules for dividing data at each node, often based on statistical measures (e.g., Gini impurity, entropy) that help in distinguishing between different defect types and severities in an efficient manner, and (4) Pruning: The process of removing parts of the tree that do not contribute significantly to decision making, to prevent overfitting and improve generalization.

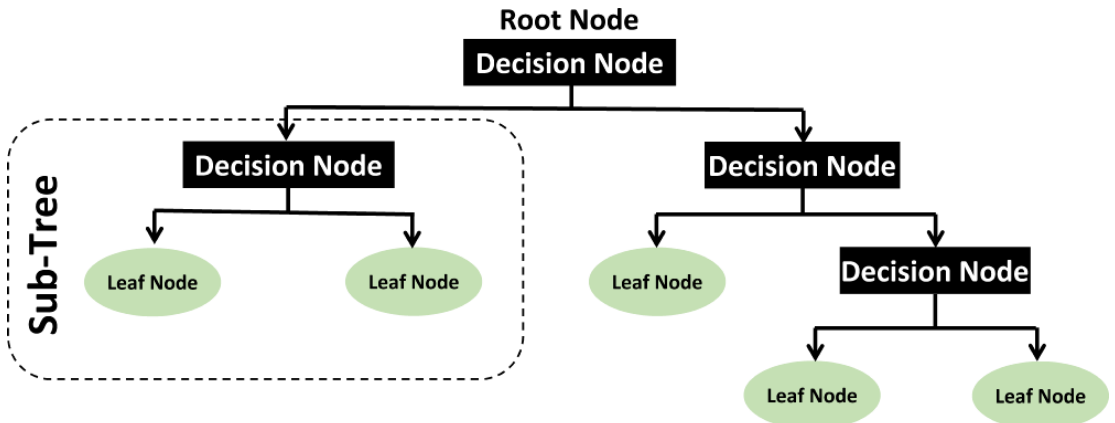

**Fig. 4:** The figure depicts the procedure of decision tree.

*b) The rationale behind the usage of the technique*

(1) Feature Selection/Importance: Decision trees excel in isolating key features for decision-making, a critical factor in wafer defect identification. They efficiently identify essential attributes signaling defects by dividing data based on various features, (2) Complex Relationships: Decision trees are well-suited for handling the nonlinear and complex relationships between wafer features and defects. They form nonlinear decision boundaries to grasp these interactions, (3) Robustness: They focus only on relevant features, ignoring insignificant ones, improving accuracy, and (4) Missing Data: Effective in handling missing data through surrogate splits or assigning values to common classes.

*c) Research Papers that have Employed the technique*

Piao et al. [32] used a decision tree ensemble and Radon transform-based features derived from raw wafer map data to recognize failure patterns and identify defect patterns in wafer maps. The final decision combines predictions from the ensemble. Chou et al. [33] developed a system using a decision tree and neural network to classify defects in chip-scale package images. The system preprocesses wafer surface images, extracting size, shape, location, and color features of defects for classification. Li et al. [34] presented a decision tree that incorporates DNNs for ADC. The decision tree utilizes defect images as the training dataset and attains an impressive classification accuracy.

*d) Case Studies and Application of the Technique*

An Intel's team [35] applied DT, to analyze extensive datasets for two main studies, as highlighted in Utlaut and Anderson's 2004 research. The first study focused on identifying wafer defects caused by radio frequency, and the second aimed to predict chip performance using early electrical testing. These investigations showed that DT might replace traditional statistical methods.

## 6. Support Vector Machine (SVM) Technique

The technique leverages a supervised learning model to differentiate between various types of wafer defects by finding the optimal hyperplane that separates data points (defects) into distinct classes with maximum margin. It processes information by mapping the wafer defect features into a high-dimensional space, where it then identifies the best separating boundary that minimizes classification errors.

*a) The Major Components of the Technique*

(1) Binary Classification of Wafer: SVMs effectively categorize wafers into defective and non-defective, learning to distinguish between them through training with known data samples, (2) Hyperplane as Decision Boundary: SVMs create hyperplanes to serve as a boundary between defective and non-defective wafers, essential for accurate defect prediction on wafer maps, (3) Maximization of Margin: By maximizing the margin between the hyperplane and the nearest data points, SVMs ensure precise and robust classification of wafers as defective or non-defective, (4) Kernel Trick for Non-Linear Boundaries: The kernel trick in SVMs transforms complex wafer data into a higher dimension for effective separation of non-linear defect patterns, enhancing defect classification accuracy, and (5) Handling Data Imbalance: SVMs address the imbalance in semiconductor data by tuning penalty parameters for each class

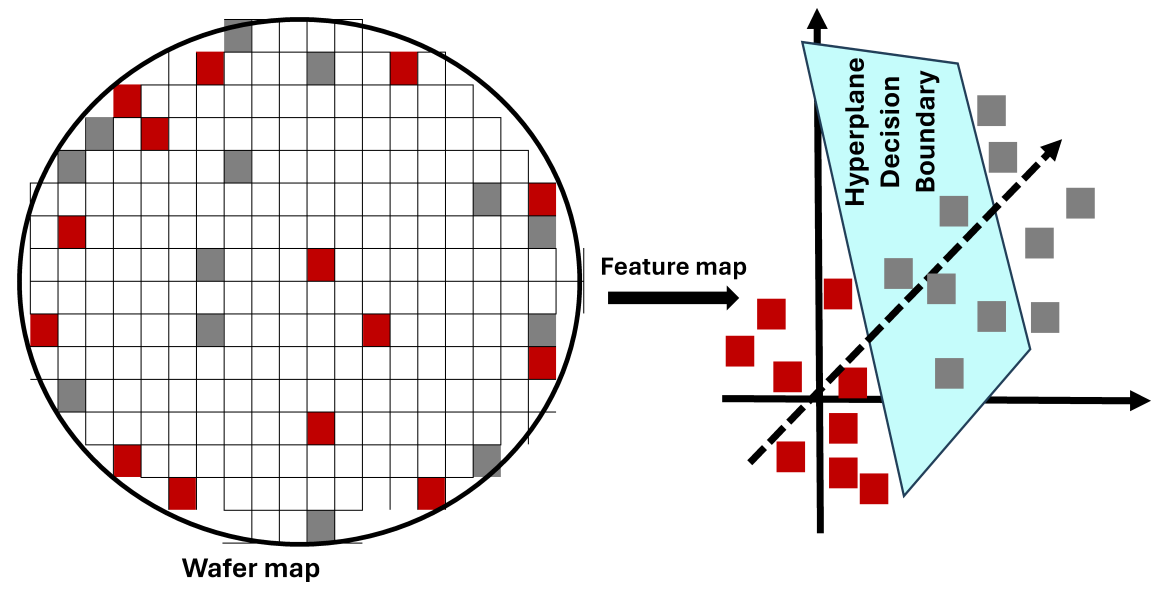

**Fig. 5:** Wafer defect pattern detection model diagram using SVM.

*b) The rationale behind the usage of the technique*

(1) Large Feature Sets: SVMs excel in managing large feature sets in wafer analysis. They adeptly learn complex patterns, crucial for detecting defects in the high-dimensional space of wafer data, (2) Robustness to Noise: SVMs are robust against noise, focusing on maximizing the margin between data classes rather than precise fitting. This minimizes outlier sensitivity, improving practical performance in tasks such as wafer defect detection, (3) Kernel Function and Non-Linear Decision Boundaries: The kernel function in SVMs enables the identification of non-linear decision boundaries by mapping data into a higher-dimensional space. This is vital for accurately detecting defects in wafer data sets, (4) Maximizing Margin: SVMs emphasize maximizing the margin between data classes, aiding in finding an optimal decision boundary. This enhances wafer defect identification by enabling generalization to unseen data, (5) Handling Imbalanced Data: SVMs effectively manage imbalanced data, common in wafer analysis. This is essential for ensuring precise defect detection.

*c) The conditions for the technique's optimal performance*

(1) Kernel Choice: For wafer defect classification, the RBF kernel is often preferred due to its flexibility in handling non-linear relationships between features, which is common in image-based data, and (2) Parameter Optimization: Cross-validation techniques such as Grid Search or Random Search can be employed to find the optimal parameter set.

*d) Research Papers that have Employed the technique*

Wu et al. [36] proposed a methodology that involves the combination of SVMs with radon-based feature extraction techniques for the purpose of predicting failure patterns. Kingma et al. [37] introduced deep generative models like the latent-feature discriminative model and generative model, employing SVM for classification and leveraging latent and continuous variables for data analysis. Li and Huang [38] applied SOM and SVM for defect spatial pattern recognition, using log odds ratio tests for systematic vs. random defect identification, SOM for clustering WBMs, and SVM for classification.

*e) Case Studies and Application of the Technique*

Baly and Hajj [39] advocate for SVMs' application in the early classification of wafers, highlighting their effectiveness in categorizing multivariate, multimodal, and inseparable data. Baly and Hajj conducted a thorough performance evaluation of SVM classifiers, using actual manufacturing data to benchmark them against current state-of-the-art methods. The core strength of SVMs is showcased in their ability to utilize multidimensional hyperplanes effectively, which skillfully segregate and categorize wafers into distinct groups of low and high yield. The study demonstrates that SVMs consistently excel over other methods.

## 7. Logistic Regression-Based Technique

The technique involves modeling the probability that a given wafer has a specific type of defect based on its features (such as patterns or anomalies detected in sensor data). The technique processes information by applying a logistic function to a linear combination of the input features to estimate the probability of each defect type. This probability is then used to classify the wafer into the most likely defect category, enabling precise identification and categorization of defects for quality control and manufacturing optimization.

*a) The Major Components of the Technique*

(1) Logistic Function: Utilizes a logistic (sigmoid) function to estimate the probability of a wafer defect based on the linear combination of selected features, (2) Optimization: Application of optimization techniques (e.g., gradient descent) to adjust the model's coefficients (weights) such that the predicted probabilities accurately reflect the likelihood of wafer defects, and (3) Decision Boundary: Establishing a threshold probability value above which wafers are classified as defective.

*b) The rationale behind the usage of the technique*

(1) Binary Classification Wafer defect classification inherently involves determining whether a wafer is defective or not, making it a binary classification problem. LR is specifically designed for binary outcomes, providing a probabilistic understanding of defect presence, which is crucial for decision-making in manufacturing processes, and (2) Handling of Imbalanced Data: Wafer defect datasets often exhibit class imbalance, where defective examples are much rarer than non-defective ones. LR can be adapted to handle such imbalances through techniques like adjusting class weights or using specialized evaluation metrics.

*c) The conditions for the technique's optimal performance*

(1) Normalization and Scaling: Given the model's sensitivity to

feature scaling, normalizing or standardizing features is essential for consistent performance and better convergence, and (2) Model Interpretability: Utilize the inherent interpretability of LR to comprehend feature importance and model decisions, aiding in defect diagnosis and resolution.

*d) Research Papers that have Employed the technique*

Saqlain et al. [40] used an ensemble method, combining logistic regression, random forest, and SVM algorithms. The success of these techniques relied on skilled feature engineering and domain expertise. Krueger et al. [41] devised a methodology using generalized linear models to predict yield in semiconductor manufacturing. Their study revealed the effectiveness of logistic regression (LR) in modeling yield based on defect data. The nested die-level LR models demonstrated superior predictive capabilities.

*e) Case Studies and Application of the Technique*

Dong et al. [42] demonstrated that logistic regression is an effective tool for modeling semiconductor yields using defect data. Their empirical studies revealed that this method is versatile, applicable to a range of device sizes and types. They highlighted logistic regression's shallow structure as a key advantage, enabling it to learn the distribution of target variables from the training data. This approach minimizes the potential biases introduced by expert experience, particularly when manufacturing processes change. Logistic regression proves to be simpler and more efficient than deep learning methods, especially with limited data.

## 8. K-Nearest Neighbor (KNN)-Based Technique

The technique classifies wafers based on the similarity of their defect characteristics to those of previously classified examples. It processes information by identifying the 'k' closest labeled data points in the feature space to the wafer under investigation, based on a chosen distance metric (e.g., Euclidean distance). The wafer is then assigned to the defect category most common among its k-nearest neighbors, leveraging the assumption that similar defect patterns are likely to belong to the same defect category.

*a) The Major Components of the Technique*

(1) Use of Distance Metrics: KNN utilizes distance metrics like Euclidean or Manhattan distance to determine the similarity between the test wafer's features and those in the training set, (2) Identification of NNs: The algorithm identifies the K closest neighbors to the test wafer. The value of K is chosen based on the complexity and variability of wafer defects, with a higher K letting a comprehensive comparison, and (3) Classification Rule: Classifies a wafer defect based on the most common defect type among its k-nearest neighbors. This step directly impacts the accuracy and reliability of defect classification.

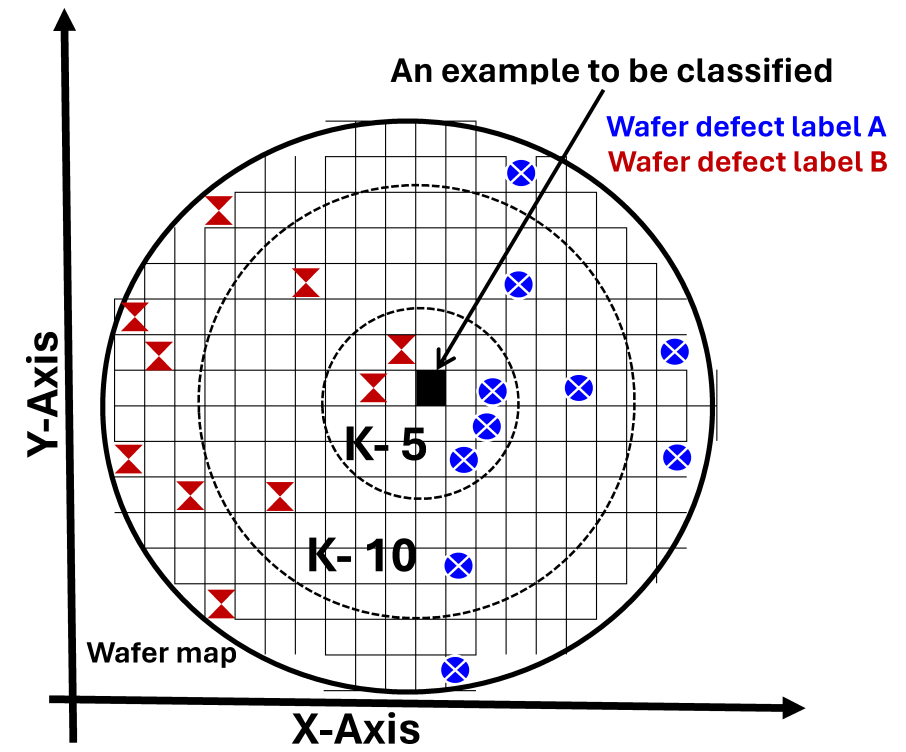

**Fig. 6:** The figure depicts the procedure of KNN.

*b) The rationale behind the usage of the technique*

(1) Feature Versatility: Wafer defect classification often involves analyzing high-dimensional data. KNN can handle this high-dimensional space effectively, making it a good choice for identifying patterns or similarities among different defect types, (2) Adaptability: KNN's effectiveness can be easily adjusted by changing the 'K' value. This adaptability allows for fine-tuning the balance between precision and recall in the classification, which is crucial in wafer defect detection, and (3) Local Decision: The algorithm bases its classification decisions on the closest examples in the feature space. This local decision-making process is particularly suited to wafer defect classification.

*c) The conditions for the technique's optimal performance*

(1) Distance Metric: Common metrics are Euclidean, Manhattan, and Minkowski. The choice depends on the data nature and wafer defect types being classified, (2) Imbalanced Data: Imbalance in classes, like one wafer defect type being more prevalent, can skew KNN results. Techniques such as oversampling less common defects or undersampling more common ones can help, (3) 'K' Value Selection: The number of 'K' impacts the result. A small 'K' may be noise-sensitive, while a large 'K' can be other class points.

*d) Research Papers that have Employed the technique*

Pan et al. [43] introduced an enhanced KNN algorithm targeting real-time detection of single-type defects, specifically scratches, to mitigate yield loss. The method begins with a skeleton extraction technique to outline scratches' main features. It then employs a clustering approach to organize these features, Kim et al. [44] developed a method using matrix factorization and KNN to categorize DRAM wafer failure patterns, illustrating machine learning's role in semiconductor quality control.

*e) Case Studies and Application of the Technique*

Yuan, T et al. [45] highlighted the efficacy of the KNN approach in wafer defect data noise reduction. This method distinguishes between global and local defects, and it compiles comprehensive data on all local defects aggregated across the wafer. It classifies these defects into clusters based on similarity clustering techniques. The approach employs a parametric model to identify and analyze the spatial patterns of these defect clusters.

## 9. Learning Vector Quantization (LVQ) Technique

The technique involves training a set of prototype vectors to represent the different categories of wafer defects. The process involves iteratively adjusting these vectors to better match the distribution of the training data, where each vector is assigned to a specific defect category. During classification, LVQ compares a new wafer's defect pattern to these prototype vectors, assigning the defect to the category of the closest matching vector, capturing the spatial and feature-based variations characteristic of wafer defects.

*a) The Major Components of the Technique*

(1) Learning Process: Through iterative comparison between input wafer maps and prototype vectors, the LVQ algorithm adjusts the prototypes to minimize classification errors. This process involves moving the prototype vectors closer to inputs that are correctly classified and away from wrongly classified ones, enhancing the model's ability to discriminate between different types of wafer defects, (2) Defect Classification: The final decision layer, where each neuron corresponds to a specific defect category. An input wafer map is classified based on the closest matching prototype vector, effectively assigning each map to a defect category, (3) Output: The output is the classification of each input wafer map into categories based on the identified defects.

*b) The rationale behind the usage of the technique*

(1) Overview: LVQ creates prototypes for defective and non-defective wafers, aiding in new wafer assessment, (2) Identifying Defects: It compares new wafers with prototypes to classify them, crucial for quality control, (3) Expert Labels: LVQ's accuracy relies on initial training with expert-labeled wafers for precise unfamiliar wafer classification, (4) Complexity: LVQ is robust against noisy data/class variations, suitable for real-world datasets.

*c) The conditions for the technique's optimal performance*

(1) Initial Prototype Selection: Choose initial prototype vectors impacts performance; they should accurately represent defects across classes, (2) Learning Rate Optimization: The learning rate should be balanced, neither too high nor too low; adaptive rates can be beneficial, (3) Algorithm Variants: The choice of LVQ variant depends on the specific wafer defect classification needs, (4) Integration with Other Techniques: Combining LVQ with other ML methods like clustering or decision trees can enhance performance, (5) Training: Proper training duration and convergence criteria are essential to avoid overfitting.

*d) Research Papers that have Employed the technique*

Chang et al. [46] introduced an LED wafer defect inspection method employing the LVQ neural network, focusing on extracting geometric and texture features from die images and their regions of interest for network training. Su et al. [47] devised a method for inspecting wafers post-sawing using learning vector quantization, achieving inspection times under one second per die, demonstrating efficiency.

### 10. Network with Self Calibrated Technique

The technique incorporates a self-calibrating mechanism to dynamically adjust the convolutional filters based on the specific characteristics of wafer defects. This self-calibration allows the network to fine-tune its feature extraction capabilities, enabling more precise identification and classification of various wafer defect types. The processing involves the network automatically learning and adjusting its parameters in response to the unique patterns of defects.

*a) The Major Components of the Technique*

(1) Convolutional Layers: Extracts features from the input wafer images through convolutional operations. These layers are adept at identifying patterns and textures associated with different types of wafer defects, (2) Self-Calibration Mechanism: Adjusts the network's parameters automatically to enhance its sensitivity to defects. This component is critical for adapting to the diverse and often subtle nature of wafer defects, (3) Pooling Layers: Reduces the spatial dimensions of the feature maps to decrease computational complexity and overfitting, (4) Fully Connected Layers: These layers interpret the features extracted by convolutional and pooling layers to classify the type of wafer defect, and (5) Output Layer: Provides the classification results, often through a softmax function that assigns probabilities to various defect categories.

*b) The rationale behind the usage of the technique*

The network with self-calibrated convolutions enhances wafer defect classification by introducing a mechanism that adaptively recalibrates feature responses, improving feature learning and the ability to capture complex defect patterns. This increases the model's robustness against input variations and ensures computational efficiency for manufacturing applications. Also, its adaptability and scalability allow for effective handling of new defect types, making it a versatile solution for maintaining quality semiconductor production.

*c) Research Papers that have Employed the technique*

Liu et al. [48] introduced a novel self-calibrated convolution that enables heterogeneous utilization of convolutional filters within a convolutional layer. They introduced an adaptive response calibration operation to encourage filters to exhibit diverse patterns. Chen et al. [49] proposed a CNN enhancement for defect detection, incorporating a multi-head attention layer for better information processing and focus on diverse input segments, utilizing self-calibrating networks.

### 11. Hopfield Neural Network (HNN)-Based Technique

The technique utilizes a content-addressable memory system for pattern recognition, making it highly effective for identifying specific patterns of defects on wafers. By iteratively updating the network state according to an energy minimization principle, HNN converges to a stable state that corresponds to a pre-learned pattern, allowing for the classification of wafer defects based on their unique patterns.

*a) The Major Components of the Technique*

(1) Neurons: Represent wafer defective features. Each neuron can be thought of as encoding a binary state related to a specific defect characteristic, (2) Weights: Encode the relationship between pairs of neurons, reflecting the correlation between different defect features. These are adjusted to capture the unique patterns of wafer defects during training, and (3) Energy Function: A measure used by the network to assess the "goodness" of a given state. The network seeks to minimize this function, guiding it towards stable states that correspond to known defect patterns.

*b) The rationale behind the usage of the technique*

(1) Energy Minimization: The operational principle of HNNs is to minimize an energy function that guides the network to converge to a stable state, which corresponds to a memory or a learned pattern. For wafer defect classification, this means the network will naturally gravitate towards known defect configurations, allowing for the reliable classification of defects, and (2) Parallel Processing: HNN operates in a parallel processing manner, which is beneficial for the rapid classification of wafer defects. Given the large volume of wafers processed in manufacturing, the ability to quickly analyze and classify defects is crucial for maintaining production efficiency and throughput.

*c) Research Papers that have Employed the technique*

Chang et al. [50] proposed an automated die inspection approach using a contextual-Hopfield neural network. The inspection is performed in multiple steps, targeting different regions, and the results are recorded on a die map. Chang et al. [51] proposed a method using a Hopfield neural network to classify wafer images by incorporating spatial information. They extended the 2-D Hopfield network to a two-layer 3-D architecture [52], enabling the detection of defective regions and integrating spatial information during pixel classification.

### 12. Adaptive Boosting (AdaBoost)-Based Technique

AdaBoost employs ensemble technique that combines multiple weak classifiers into a strong classifier to improve the accuracy of defect identification on wafers. It iteratively adjusts the weights of incorrectly classified instances so that subsequent classifiers focus more on difficult cases, effectively enhancing the model's sensitivity to subtle and complex defect patterns. This process allows AdaBoost to efficiently classify wafer defects by progressively refining its decision boundaries.

*a) The Major Components of the Technique*

(1) Weak Learners: Utilizes simple models, often decision trees, that can distinguish between defect and non-defect classes or classify types of defects on wafers, (2) Weighted Data Points: Initially assigns equal weights to all training samples. Subsequently adjusts these weights to emphasize the importance of samples misclassified by previous learners, ensuring the model learns from its mistakes, (3) Iterative Learning: Sequentially applies weak learners to increasingly weighted datasets, focusing on challenging samples that previous learners misclassified, and (4) Aggregation of Learners: Combines the output of multiple weak learners into a strong classifier by weighting their predictions based on their accuracy, effectively improving the model's ability to detect and classify wafer defects accurately.

*b) The rationale behind the usage of the technique*

(1) Enhancing Detection: Wafer defects can vary significantly in size, shape, and appearance, with some being extremely subtle and easy to overlook. AdaBoost improves the model's sensitivity to such variations by combining multiple weak classifiers, each potentially adept at recognizing different types of defects. This ensemble approach enhances the overall ability of the system to identify a wider range of defect types, including those that are not easily detectable by individual classifiers, and (2) Adaptability to Feature Diversity: In wafer defect classification, the relevant features for identifying defects can be diverse, ranging from geometric patterns to textural and spectral characteristics. AdaBoost's flexibility in integrating different types of weak learners allows it to exploit this feature diversity effectively.

*c) Research Papers that have Employed the technique*

Zuo et al. [53] improved wafer testing accuracy and reduced false failures using AdaBoost Tree, effectively handling data imbalance and better identifying critical defects. Lee et al. [54] improved wafer defect classification with the AdaBoost classifier by extracting features from local wafer image regions.

## B. Multi-Label Classification Sub-Category

### 1. Generative Adversarial Network (GAN)Technique

The technique involves a dual-network architecture, comprising a generator and a discriminator, working in opposition to enhance the classification of various defect types on wafers. The generator attempts to create synthetic wafer images with defects, while the discriminator evaluates these images against real defect data to improve its ability to classify multiple defect types. Through this adversarial process, the system iteratively refines its capability to distinguish between different defect categories, improving the accuracy and reliability of wafer defect classification by learning from complex, multi-label datasets.

*a) The Major Components of the Technique*

(1) Generator: Creates synthetic images of wafers, incorporating multiple types of defects, mimicking real-world variations. It is tailored to simulate a diverse range of wafer defect patterns, (2) Discriminator: Analyzes images to determine if they are real or synthetic, simultaneously classifying the types of defects present. It is fine-tuned for high precision in distinguishing between various defect types on wafers, crucial for the accurate identification and classification of complex, multi-label defects, (3) Adversarial Training: It Involves the generator and discriminator in a competitive training process to progressively improve wafer defect classification performance. It can be optimized, focusing on enhancing the system's ability to detect and classify multiple, subtle defect types on wafers, essential for quality control and yield improvement.

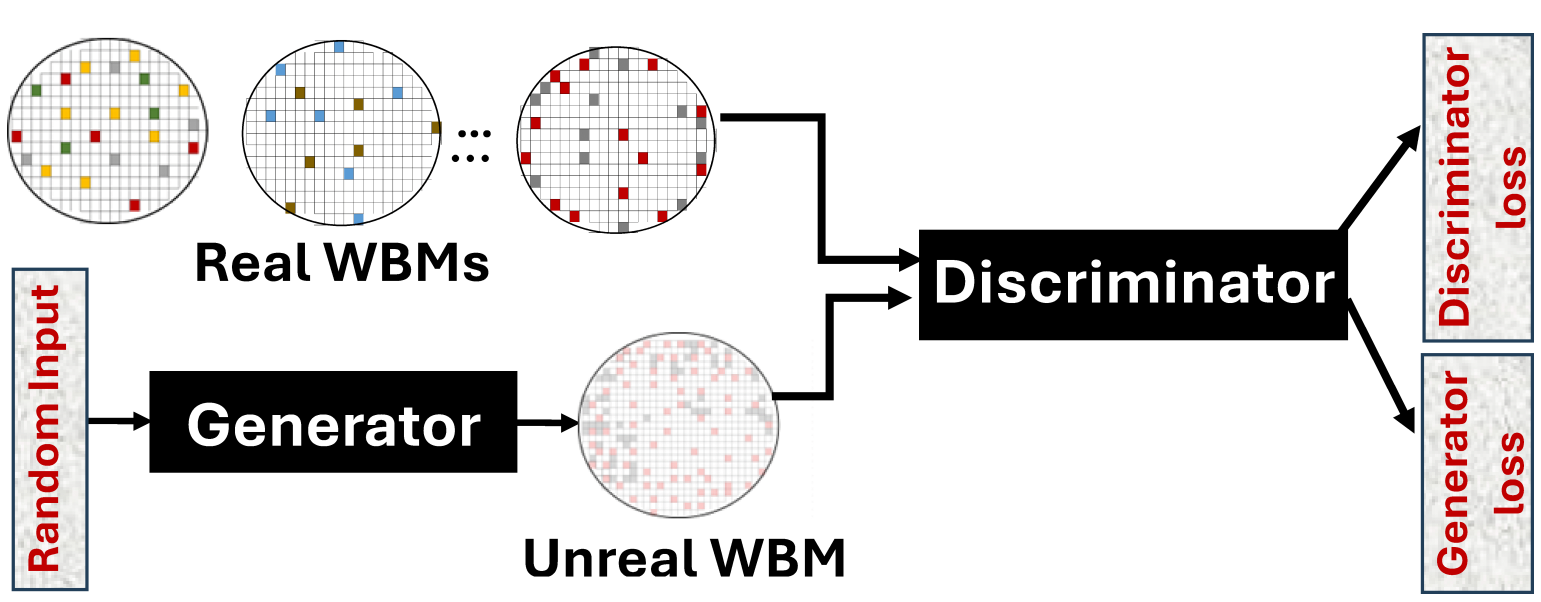

**Fig. 7:** Wafer defect pattern detection model diagram using GAN.

*b) The rationale behind the usage of the technique*

(1) Overcoming Data Scarcity: Semiconductor manufacturing often deals with highly proprietary and sensitive data, leading to limited availability of defect samples for training. GANs can generate realistic, synthetic wafer images with multiple defect types, augmenting the dataset and enabling effective training even in data-constrained environments, (2) Handling Complex, Multi-Label Scenarios: Wafer defects can be numerous, varied, and often occur simultaneously, requiring a system capable of identifying multiple defect types within a single image. GANs, through their adversarial training process, learn to produce and refine synthetic images that closely resemble real wafers with multiple defects, thereby enhancing the discriminator's ability to accurately classify complex, multi-label defects, (3) Improving Detection Accuracy: The adversarial process inherent in GANs continuously challenges the discriminator to improve its ability to distinguish between real and synthetic images and identify the types of defects present.

*c) Research Papers that have Employed the technique*

Shim et al. [55] introduced a technique for training CNN that involves the use of multi-label training wafer maps for precise classification of mixed-defect wafer maps. This method incorporates three key elements to create synthetic wafer maps from a multi-label training dataset: mixup, random rotation, and noise filtering. The mixup component is used to merge single-defect wafer maps, thereby generating synthetic maps with mixed defects. Byun and Baek [56] developed a deep convolutional GAN that synthesizes wafer maps to generate multi-label defects by combining single-type patterns through pixel-wise summation. Lee et al. [57] introduced a semi-supervised, multi-label learning approach for categorizing WBMs based on various defect patterns. They utilized Generative Adversarial Networks to effectively leverage both labeled and unlabeled data. Their methodology frames the classification of mixed-type defect patterns as a multi-label classification issue. By identifying the presence or absence of individual distinct patterns, they categorized WBMs into 16 different classes.

*d) Case Studies and Application of the Technique*

Intel [58] experimented with GenAI models, including GANs and diffusion models, showing success. Utilizing SPICE parameters in device simulation, these models predict the electrical characteristics (ETEST metrics) of devices, accurately forecasting the ETEST metric distribution. Accurate predictions of circuit yield distribution enable optimization at the design stage, leading to cost savings, shorter development times, and increased yields, benefiting foundries and design teams that incorporate these models into their processes.

### 2. Support Vector Machine (SVM) Technique

The technique involves categorizing wafer defects into multiple labels simultaneously, leveraging SVM's capability for handling complex, high-dimensional data. This approach uses SVM's margin-based classification to distinguish between various types of defects on wafers, enhancing the precision and efficiency of defect identification. By training the SVM model with labeled wafer images showcasing different defects, the technique enables the classification of new images into multiple defect categories, improving manufacturing quality control.

*a) The Major Components of the Technique*

(1) SVM Classifier: For multi-label classification, either multiple binary SVMs are trained for each defect type (one-vs-all approach) or a single SVM is adapted to output multiple labels, and (2) Multi-label Strategy: This can involve modifying SVM to directly handle multi-label data or an ensemble of multiple SVM classifiers, each responsible for a subset of labels.

*b) The rationale behind the usage of the technique*

(1) Strength in High-Dimensional Spaces: SVM is particularly effective in high-dimensional spaces, which is common in wafer defect classification due to the complex and detailed data captured by imaging and sensor technologies, (2) Marginal Optimization: SVM focuses on optimizing the decision boundary margin, making it well-suited for distinguishing between different defect types, even when the differences are subtle.

*c) Research Papers that have Employed the technique*

Chao and Tong [59] introduced an innovative recognition system that employs multi-label support vector machines along with a newly developed defect cluster index. This system is designed to recognize wafer defect patterns with high efficiency and accuracy. Fan et al. [60] introduced a multi-label approach for identifying wafer defect patterns, enabling the recognition of multiple defect patterns simultaneously using SVM and the Ordering Point To Identify the Cluster Structure (OPTICS) algorithm.

*d) Case Studies and Application of the Technique*

A Massachusetts Institute of Technology (MIT) researcher [60] applied a statistical prediction model to optimize operations at two manufacturing sites. During its pilot phase, the model demonstrated the potential of SVM and random forest classifiers in refining both the wafer sort process and wafer defect identification at the wafer and die levels.

*e) The conditions for the technique's optimal performance*

(1) Implementing an appropriate strategy for handling multi-label data, such as one-vs-rest (OvR) or one-vs-one (OvO), to adapt SVM for multi-label tasks, and (2) SVM can be computationally intensive. Adequate computational resources are necessary, particularly for handling large datasets.

## III. Agent-Based Classification Category

### A. Single-Agent Classification Sub-Category

### 1. Hidden Markov Tree (HMT)-Based Technique

The technique leverages the hierarchical and sequential nature of data to model the spatial dependencies among wafer defect patterns. This technique processes information by employing a tree-structured graphical model where nodes represent defective states that are not directly observable (hidden states), and the edges encode the conditional dependencies between these states. By analyzing the sequential data through probabilistic transitions among states, it classifies wafer defects by recognizing underlying patterns and correlations in the spatial arrangement of defects.

*a) The Major Components of the Technique*

(1) State Space: The state space in an HMT for wafer defect classification defines the possible conditions or states that each section of the wafer can be in, such as different types of defects or no defect, (2) Transition Probabilities: It quantifies the probability of a defect type changing to another type or remaining the same from one area to another, and (3) Tree Structure: The HMT tree structure represents the hierarchical relationship between different parts of the wafer, enabling the model to analyze defects at various scales and resolutions. Fig. 8 depicts the procedure of HMT.

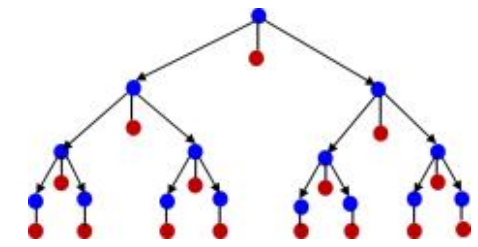

**Fig. 8:** The figure depicts the process of Hidden Markov Tree

*b) The rationale behind the usage of the technique*

(1) Modeling Complex Spatial Relationships: Wafer defect patterns exhibit complex spatial relationships that can be indicative of the underlying manufacturing issues. The HMT is particularly well-suited for modeling these relationships because it extends the Hidden Markov Model to tree structures, allowing for the modeling of hierarchical and multi-scale spatial dependencies among defects, (2) Multi-Scale Analysis: The HMT's inherent structure supports multi-scale analysis, enabling it to classify defects accurately across different scales, and (3) Agent-Based Adaptability: Incorporating the HMT within an agent-based framework enhances the system's adaptability and scalability. Each agent, equipped with HMT capabilities, can focus on different aspects of the wafer inspection process (e.g., detecting types of defects or analyzing regions of the wafer). This modular allows for parallel processing and easy scalability.

*c) Research Papers that have Employed the technique*

Zhou [62] presented an innovative method for online detection and recognition of wafer surface defects, leveraging hidden Markov dynamic integration to extract features and construct Hidden Markov Models for adaptive recognition, validated by its effectiveness on the WM-811K database. Similarly, Chen et al. [63] introduced a method using a growing wavelet-based hidden Markov tree (gHMT) for the automated identification of defects in spatial wafer maps, capable of accurately characterizing and locating defect regions by analyzing statistical properties and patterns in defect spatial data.

### 2. Mean Shift-Based Technique

The technique identifies clusters of defect data points on a wafer map by iteratively shifting towards the regions of maximum data density. This process leverages the spatial information of defects, enabling the algorithm to dynamically adjust the window size for each cluster based on the density gradient, thereby accurately segregating different types of wafer defects. This density-based approach allows for the precise identification and classification of wafer defects without the need for specifying the number of clusters in advance.

*a) The Major Components of the Technique*

(1) Mean Shift Clustering: It identifies centroids (mean points) of data points that are densely packed together in the feature space, effectively grouping similar features (indicative of specific wafer defect types) into clusters, (2) Agent-Based Modeling: Agents are defined with specific roles or behaviors for navigating through the clustered feature space to classify wafer defects. Each agent may represent a different defect classification rule or strategy, dynamically interacting with the clusters identified by the Mean Shift algorithm to assign defect types, and (3) Defect Classification: This component involves the actual classification of defects based on the interactions between agents and the clustered feature space. The outcome is a labeled wafer map, where each defect is classified into predefined categories (e.g., scratches, pits, particles).

*b) The rationale behind the usage of the technique*

(1) Density-Based Clustering for Anomaly Detection: Mean Shift operates by identifying dense regions of data points in a feature space, which aligns well with the nature of wafer defects. Defects on wafers tend to form clusters, due to specific process malfunctions/material inconsistencies. Mean Shift can effectively detect these dense clusters without prior knowledge of the number of clusters, (2) Robustness to Noise and Outliers: Mean Shift's reliance on local density estimates makes it inherently robust to noise and outliers, and (3) Agent-Based Model Integration: Each agent (e.g., a segment of the wafer) can independently apply the Mean Shift algorithm to locally detect defects. This decentralized approach allows for scalable and parallel processing of wafer inspection data, leveraging Mean Shift's capability to operate without global information about data distribution.

*c) Research Papers that have Employed the technique*

Tsai and Luo [64] introduced a machine vision method using mean shift and gradient direction entropy to identify defects in multi-crystalline solar wafers, creating a feature space combining pixel coordinates and entropy. Bousetta and Cross [65] examined how wafer sampling adjustments, based on metrics like normalized mean shift, variance ratio, and excursion frequency, can optimize monitoring and response to changes in wafer defect distributions.

### 3. Density-Based Technique

The technique revolves around leveraging localized density variations on wafers to identify and classify defects. By deploying agent-based models, this technique assigns agents to navigate through wafer surface data, utilizing statistical analysis to detect anomalies indicative of various defect types. The process involves agents evaluating the density of points (representing potential defects) within their vicinity, allowing for the classification of defects based on predefined density thresholds and patterns.

*a) The Major Components of the Technique*

(1) Classification Agents: Classification agents analyze the data collected by sensing agents to identify specific types of defects. They employ machine learning algorithms or rule-based systems to categorize defects into predefined classes such as scratches, particles, or voids, based on their characteristics, (2) Coordination Mechanism: This component ensures effective communication and coordination among agents. It manages the flow of information between sensing, classification, and decision-making agents, ensuring that data is accurately shared and actions are properly synchronized across the system.

*b) The rationale behind the usage of the technique*

Localized Defect Detection: Density metrics in agent-based models allow for the precise detection of defects at a very localized level. Agents can be programmed to monitor specific regions of a wafer, analyzing density variations to identify potential defects. This localized approach ensures that even small areas with high defect concentrations are detected, improving the defect detection capability, and (2) Learning from Patterns: Over time, agents can learn from the density patterns of defects, improving their classification algorithms. This learning process enables the system to become increasingly effective at identifying/classifying defects.

*c) Research Papers that have Employed the technique*

Jin et al. [66] developed a defect detection framework for Wire Bonding Machines using clustering, starting with the identification of faulty and edge die for DBSCAN clustering. Cheng et al. [67] integrated automatic test equipment data with the NXP dataset, using DBSCAN and image processing to distinguish between testing and foundry defects. Tan and Lau [68] proposed a method to automate wafer map extraction with DBSCAN, aiming to replace the Manual Visual Inspection method while assessing the optimal size for clustered signatures. Koo and Hwang [69] introduced a two-step defect pattern analysis using density-based clustering, first identifying abnormal wafer maps statistically, then clustering defect patterns.

## B. Multi-Agent Classification Sub-Category

### 1. Hierarchical Agglomerative-Based Technique

The technique utilizes a bottom-up approach, starting with each wafer defect as its own cluster and iteratively merging them into larger clusters based on similarity measures. This process relies on defining a precise metric or distance between defects to evaluate similarity, incorporating factors like wafer defect shape, size, and distribution patterns on the wafer. The technique focuses on gradually building a hierarchy of defect classes, enabling categorization and analysis of wafer defects.

*a) The Major Components of the Technique*

(1) Similarity Measurement: Determines the closeness between defects using metrics tailored to wafer defect patterns, such as Euclidean distance, (2) Clustering Algorithm: Employs a hierarchical agglomerative clustering approach, initially treating each defect as a separate cluster and iteratively merging clusters based on similarity, specifically designed to handle the diversity and complexity of wafer defect types, (3) Multi-Agent System: Incorporates multiple agents, each specializing in different aspects of the defect classification process, such as feature extraction, similarity measurement, or cluster merging, to enhance the accuracy and efficiency of the classification, and (4) Hierarchical

Structure: Organizes defect classes into a hierarchical tree, facilitating a detailed/scalable classification that accommodates varying levels of defect granularity.

*b) The rationale behind the usage of the technique*

(1) Adaptability to New Defects: HAC's hierarchical nature allows for incremental updates to the classification scheme without needing to retrain the model from scratch, (2) Insightful Defect Hierarchy: HAC organizes defects into a tree-like hierarchical structure that mirrors the natural grouping of defect types, from very general to very specific. This structure can reveal insights into the underlying causes of defects, their relationships, and their severities, aiding in root cause analysis and process improvement efforts, and (3) Decentralized Analysis: In a multi-agent system, different agents can perform HAC on subsets of data, and their results can be aggregated to form a comprehensive defect classification. This leverages the hierarchical structure of HAC to merge individual analyses into a global overview efficiently.

*c) Research Papers that have Employed the technique*

Yu and Liu [70] proposed a Hierarchical Agglomerative multigranularity generative adversarial network, focusing on enhancing wafer maps through multi-agent network interaction, consisting of an auxiliary feature extractor, a generator, and a discriminator. Wang et al. [71] integrated Hierarchical Agglomerative and fuzzy C-means algorithms to distinguish and classify different types of defect patterns.

## 2. Statistical Hierarchy-Based Technique

The technique primarily leverages hierarchical statistical modeling and multi-agent systems to efficiently classify wafer defects. It processes information by first segmenting the wafer surface into distinct regions and employing multiple, specialized agents that analyze these regions for various defect types. Each agent applies statistical models to evaluate the likelihood of specific defects, with the hierarchy enabling prioritization and aggregation of findings to improve classification accuracy.

*a) The Major Components of the Technique*

Hierarchical Statistical Models: Utilized to aggregate and interpret data from multiple agents, employing a tiered approach to refine defect classification accuracy through successive levels of analysis, (2) Data Segmentation: It divides the wafer surface into segments, allowing agents to efficiently process and analyze data by focusing on smaller, manageable areas, (3) Classification Algorithms: Algorithms that integrate inputs from multi-agent systems and hierarchical models to accurately classify the type and severity of wafer defects.

*b) The rationale behind the usage of the technique*

(1) Complexity of Wafer Defects: The statistical hierarchy allows for the modeling of complex wafer defects by breaking down defect types into hierarchical categories, enabling more nuanced analysis and classification, and (2) Need for Precision: The statistical hierarchy aids in minimizing false positives and false negatives by providing a structured framework for defect classification, allowing for the application of specific statistical tests and models that can more accurately distinguish between defect types and severities.

*c) Research Papers that have Employed the technique*

Zhang and Wang [72] proposed a multi-agent collaborative system for semiconductor manufacturing with three levels: system, machine, and material. The system layer aims to maximize processing profit, while the machine layer's goal is to select the winning bids. Mönch et al. [73] detail a prototype of a multi-agent system (MAS) structured hierarchically, aimed at controlling the production and inspection of wafer fabrication. They propose a three-tiered hierarchy within the MAS.

## 3. K-Means-Based Classification ML Sub-Technique

The technique employs an iterative clustering algorithm to categorize wafer defects into distinct groups based on their features, such as size, shape, and location. This technique processes information by first initializing centroids randomly, then assigning each defect to the nearest centroid based on Euclidean distance or another relevant metric, thereby forming clusters. It iteratively updates the centroids by calculating the mean of the points within each cluster until convergence, effectively grouping defects with similar characteristics for efficient identification.

*a) The Major Components of the Technique*

Distance Metric: Employs a suitable distance metric (e.g., Euclidean distance) tailored to wafer defect characteristics, enabling the precise measurement of similarity between defects and centroids, (2) Clustering Algorithm: Applies the K-Means algorithm to iteratively assign defects to the nearest cluster based on their features, optimizing the placement of centroids to form cohesive groups that reflect underlying wafer defect patterns, and (3) Convergence: Determines the point at which the algorithm stops iterating, such as when centroid positions stabilize, ensuring efficient classification without unnecessary computation.

*b) The rationale behind the usage of the technique*

(1) Pattern Recognition in High-Dimensional Data: K-Means clustering excels in identifying patterns within high-dimensional datasets by grouping data points (i.e., instances of potential defects) based on feature similarity. This capability is crucial for detecting and classifying the myriad defect types that can occur on a wafer, (2) Flexibility in Defining Defect Categories: K-Means provides the flexibility to identify clusters of defect instances that exhibit similar characteristics, thereby facilitating the categorization of defects into meaningful groups. This dynamic categorization can be particularly useful for identifying new or evolving defect types over time, and (3) Integration with Multi-Agent Systems: K-Means can serve as an initial clustering step that groups similar defects together. Subsequent agents can then perform more granular analysis or classification within these clusters, leveraging additional domain knowledge or specialized machine learning models tailored to specific defect types.

*c) Research Papers that have Employed the technique*

Jubair et al. [74] introduced a parallel clustering method that merges the principles of K-means with Multi-Agent System (MAS) algorithms, termed Multi-K-means (MK-means). This approach aims to maintain the integrity of the dataset while enhancing the clustering accuracy. It involves calculating the cluster centers for each partition, merging them, and subsequently performing clustering. Jubair et al. [75] introduced a novel parallel clustering algorithm that integrates Multi-Agent Systems and the K-means algorithm, named Multi-agent-K-means (MK-means). The MK-means algorithm employs a separate activation agent for clustering, taking into account a distinct subset of features for each agent. The objective is to enhance clustering accuracy while ensuring the dataset remains unchanged.

# IV. Type-Based Classification Category

## A. Single-Type Classification Sub-Category

### 1. Convolutional Generative Adversarial Network (CGAN)

The technique leverages the adversarial relationship between a generative network, which produces synthetic wafer defect images, and a discriminative network, which evaluates whether images are real defects or synthetic. It continually improves the generative model's output through feedback from the discriminative model, leading to more precise wafer defect recognition and classification. The key concept revolves around using convolutional neural networks within this adversarial framework to efficiently process and learn from the spatial hierarchy of features in wafer images, enabling differentiation of defect types.

*a) The Major Components of the Technique*

(1) <u>Generator:</u> Utilizes a series of deconvolutional layers to upscale latent space vectors into detailed images. Each layer progressively increases the resolution, focusing on generating the complex patterns of defects seen in wafers, and (2) <u>Discriminator</u>: Comprises convolutional layers that progressively downsample the input wafer images, extracting features critical for wafer defect classification. This component identifies a wide range of wafer defect types. Fig. 9 depicts the procedure of CGAN.

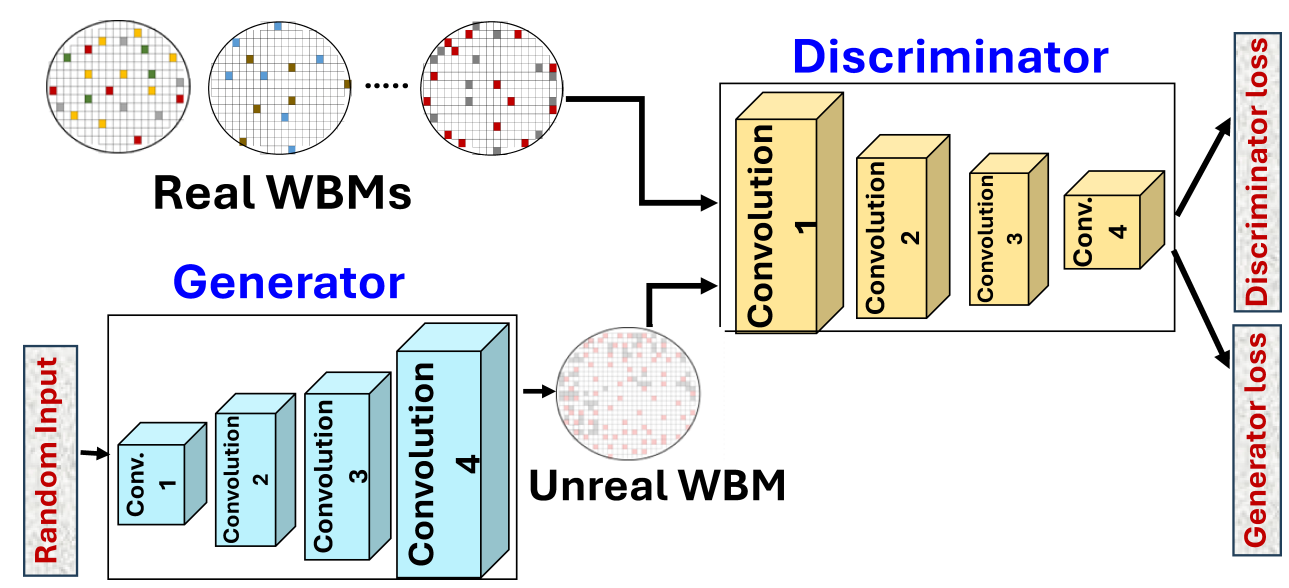

**Fig. 9:** Wafer defect pattern detection model diagram using CGAN.

*b) The rationale behind the usage of the technique*

(1) Improved Generalization Capability: CGANs, through their generative capabilities, produce a wide variety of wafer defect images, which helps the model learn to recognize and classify a broader spectrum of defect types more accurately, and (2) Adversarial Learning Process: The adversarial learning process inherent in CGANs, where the generative model continuously strives to improve the quality of synthetic images to fool the discriminative model, inherently leads to the generation of highly realistic wafer defect images. This process not only enhances the training dataset but also pushes the discriminative model to develop more refined classification capabilities.

*c) Research Papers that have Employed the technique*

Byun and Baek [76] developed a method using a deep convolutional generative adversarial network (DCGAN) for creating single-type wafer maps, incorporating pixel-wise addition and thresholding to maintain binary pixel characteristics. Park and You [77] proposed a DCGAN-based data augmentation technique to improve CNN classifier by generating varied defect patterns, introducing a quantitative index for evaluating augmentation effectiveness and a masking process for image refinement.

*d) Case Studies and Application of the Technique*

Samsung Electronics [78] has adopted Adversarial Training in semiconductor manufacturing. This process involves a computer-readable medium with a program for an image generation model. When activated, it guides a processor through complex steps, starting with inputting semiconductor die samples into a generator network to create a detailed wafer map. This map is then analyzed by a discriminator network within the model, which identifies and classifies defects.

### 2. Convolutional Neural Network Technique

The technique efficiently identifies specific types of defects on wafers. It processes information through a series of convolutional layers that extract and learn hierarchical feature representations from raw wafer images, effectively capturing spatial relationships and patterns indicative of defects.

*a) The Major Components of the Technique*

<u>Activation Functions</u>: ReLU (Rectified Linear Unit) introduces non-linearity after each convolution operation, allowing the network to learn complex patterns in the data relevant to wafer defect characteristics, (2) <u>Pooling Layers</u>: Pooling reduces the dimensionality of the feature maps, retaining the most essential information, (3) <u>Fully Connected Layers</u>: These layers interpret the feature representations learned by convolutional and pooling layers, mapping them to specific wafer defect types in the classification layer, and (4) <u>Output Layer</u>: The softmax function in the layer assigns probabilities to each defect class.

*b) The rationale behind the usage of the technique*

(1) <u>Spatial Hierarchy</u>: CNNs employ a hierarchical structure of convolutional layers that progressively extract and combine features from different levels of abstraction. This approach matches well with the nature of wafer defects, enabling CNNs to identify complex defect patterns effectively, and (2) <u>Translation Invariance</u>: One of the hallmark properties of CNNs is their ability to achieve translation invariance. In the context of wafer defect classification, this means that a CNN can recognize a defect regardless of its position on the wafer. This property is crucial because defects can occur anywhere on a wafer's surface.

*c) Research Papers that have Employed the technique*

Luo et al. [79] presented CWDR-Net, a CNN-based framework for wafer defect identification, utilizing the MVDFE module for enhanced, noise-resistant feature extraction and a defect type-specific, attention-driven classifier. This framework can selectively extract information from the defect pattern and class-specifically recognize each basic single-type defect. Chiu and Chen [80] applied rotational data augmentation and copy-paste methods alongside Mask R-CNN for accurate single-type and mixed-type wafer defect classification. Cheon et al. [81] demonstrated the efficacy of integrating CNN and k-NN for classifying single-defect patterns, including the detection of unknown types and enhancing model accuracy through a specific four-layer CNN and k-NN combination.

### 3. K-Means/C-Means-Based Technique

The technique involves clustering based on the similarity of wafer defect patterns. The algorithm categorizes wafer defects into clusters (or classes) by iteratively minimizing the variance within clusters and maximizing the variance between clusters.

*a) The Major Components of the Technique*

(1) <u>Clustering Algorithm</u>: Applies K-Means (hard clustering) or C-Means (fuzzy clustering) to group wafer defects into clusters based on feature similarity, optimizing for intra-cluster similarity and inter-cluster dissimilarity, (2) <u>Iterative Optimization</u>:

Iteratively updates cluster centers and reassigns defects to clusters to minimize within-cluster variances until convergence or a maximum number of iterations is reached, and (3) Cluster Assignment: Assigns each wafer defect to the cluster with the closest center (in K-Means) or calculates membership grades for each cluster (in C-Means).

*b) The rationale behind the usage of the technique*

(1) Facilitating Root Cause Analysis: By grouping similar defects together, K-Means and C-Means facilitate root cause analysis and corrective action, and (2) Dimensionality Reduction: Both algorithms can be combined with feature extraction and dimensionality reduction techniques to improve classification performance. By identifying and focusing on the most relevant features of the defects, these algorithms can more effectively cluster similar defect types, improving accuracy.

*c) Research Papers that have Employed the technique*

Chen et al. [82] improved image segmentation's noise resilience with enhanced k-means clustering and morphological filtering, achieving superior noise reduction. Pugazhenthi and Singhai [83] created an automated centroid-clustering image segmentation algorithm using k-means, highlighting its difficulty in stable wafer centroid detection due to grayscale defects. Horng and Hsiao [84] proposed a fuzzy clustering decision tree method for classifying large datasets with continuous attributes, leveraging hierarchical clustering for precise fuzzy partitioning.

## B. Multi-Type Classification Sub-Category

### 1. CNN for Multi-Type Defect Classification Technique

The technique utilizes layered architecture to extract features automatically and hierarchically from wafer images at multiple scales, enabling the identification and classification of various defect types with high precision. By employing convolutional layers to process spatial information and pooling layers to reduce dimensionality, CNN efficiently learns defect-specific patterns directly from raw wafer images.

*a) The Major Components of the Technique*

(1) Convolutional Layers: Applies filters to the input images to detect specific features like edges, textures, or patterns indicative of wafer defects, (2) Activation Function: Utilizes functions like ReLU to introduce non-linear properties, helping the network learn complex patterns in wafer defect classification, (3) Pooling Layers: Reduces the spatial size of the extracted features to decrease computational load and overfitting risk, focusing on the most relevant features for defect detection, and (4) Fully Connected Layers: After feature extraction and reduction, these layers aggregate the learned features to make a final classification decision on the type of wafer defect.

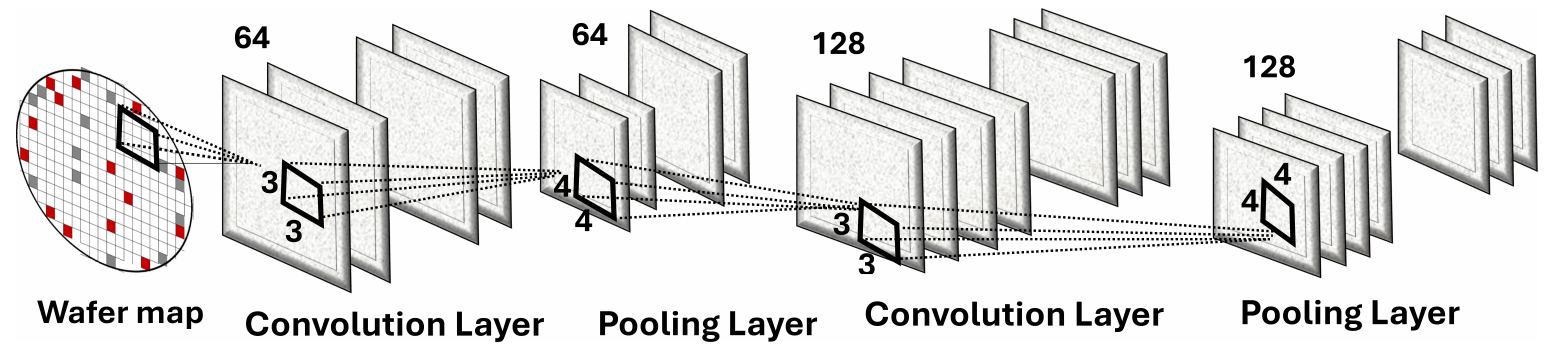

**Fig. 10:** CNN for multi-type model for wafer defect pattern detection.

*b) The rationale behind the usage of the technique*

(1) Accuracy: The layered architecture of CNNs enables the detailed analysis of wafer images. This structure allows for the precise localization and identification of wafer defects, and (2) Reduced False Negatives: The feature learning capability of CNNs minimizes the occurrence of false positives and negatives.

*c) The Conditions for the Technique's Optimal Performance*

(1) Attention Mechanisms: Implement attention mechanisms to help the network focus on areas most indicative of defects, (2) Hybrid Models: Consider integrating CNNs with other models like RNNs for handling sequences of wafer inspection images, if temporal data is available, (3) Transfer Learning: Use pretrained networks on similar tasks to jump-start the learning process, adapting these models to the specific nuances of wafer defects.

*d) Research Papers that have Employed the technique*

Battol et al. [85] developed an advanced CNN with attention mechanisms for mixed-type wafer defect classification, focusing on critical areas through spatial attention and multiple channels, employing a focal loss function and a Global Average Pooling layer. Liu and Tang [86] introduced a triplet CNN model approach for mixed-type wafer defect classification, using weakly supervised learning on imprecisely labeled datasets. Lee et al. [87] introduced SS-AIR, a semi-supervised approach for mixed-type wafer defect classification and location, leveraging CNNs with both labeled and unlabeled data, and SVM classifiers. Wei and Wang [88] presented MSF-Trans, blending Multi-Scale Information Fusion Transformer, CNNs, and transformers for mixed-type wafer defect classification and global context capture.

*e) Case Studies and Application of the Technique*

Kyeong and Kim [89] employed CNNs to classify WBMs with mixed-type defect patterns, eliminating the need for pre-removing random defects or clustering systematic defects. They applied multi-label classification by using separate CNN models for each label. Their study demonstrated CNNs' robustness against global random defects, achieving notably good accuracy, outperforming other methods in comparison. They used separate models for each defect type, showing CNN's superiority in handling mixed and numerous global random defects.

### 2. Deep Neural Network (DNN) Technique

The technique processes information through multiple layers of neurons, each layer capable of recognizing increasingly abstract features of the wafer defects, from simple edges and textures in the initial layers to complex defect structures in deeper layers. This approach enables accurate classification of multiple wafer defect types by capturing the intricate spatial relationships and variations inherent to each defect category.

*a) The rationale behind the usage of the technique*

Complexity of Defects: DNNs are adept at discerning the patterns of wide array of defects, enabling them to identify and classify a broad spectrum of wafer defect types accurately, (2) Precision and Requirements: DNNs, through their deep layered structures, have the capability to model the intricate details and variations of wafer defects, and (3) New Defect Types: DNNs can be retrained or fine-tuned with additional data to accommodate new defect patterns.

*b) Research Papers that have Employed the technique*

Li et al. [90] developed an ADC approach leveraging DNNs with a decision tree for superior image classification and a self-learning component for retraining on low-confidence "Unknown" cases, avoiding full system refreshes. Saqlain et al. [91] proposed a voting ensemble classifier for detecting wafer map defects, merging various feature types and classifiers (DNNs, logistic regression, random forests) for enhanced accuracy.

## V. OBSERVATIONAL ANALYSIS

In this section, we scrutinize the various machine learning classification strategies presented in this survey, all designed for the detection of defective patterns in wafer maps. We evaluate each technique using the following four criteria: Complexity, Performance, Robustness, and Limitations. The assessments can be found in Table 1 for the different techniques.

**TABLE 1:** EVALUATING EACH MACHINE LEARNING WAFER DEFECT CLASSIFICATION TECHNIQUE IN TERMS OF FOUR CRITERIA

| | **Complexity** | **Performance** | **Robustness** | **Limitations** |
|---|---|---|---|---|
| **CNNs** | CNNs entail a sophisticated architecture designed to learn spatial hierarchies of features from images of wafers. This complexity arises from multiple layers, which work together to detect patterns indicative of defects. The complexity is justified by the need to capture a wide variety of defect types and sizes. | CNNs have demonstrated superior performance wafer defect classification. They excel at identifying intricate patterns and anomalies on the wafer surface with high accuracy and speed. When optimally configured, CNNs can achieve impressive wafer defect classification accuracy, reducing the rate of false positives/negatives. | CNNs exhibit a high degree of robustness in wafer defect classification, particularly in handling variations in defect appearance due to different manufacturing processes or environmental conditions. Their ability to learn feature representations automatically enables them to adapt to new defect types or changes in defect characteristics over time | (1) Their performance heavily relies on the availability of large, annotated wafer maps for training, which can be resource-intensive, (2) CNNs is considered "black boxes" due to their complex structures, making it difficult to understand the reasoning behind specific classifications or predictions, (3) CNNs may struggle to adapt to new or unseen types of wafer defects without retraining or fine-tuning. |
| **Residual Neural Network** | ResNets feature skip connections that let inputs bypass layers, addressing the vanishing gradient problem in deep neural networks. This complexity aids in learning nuanced wafer defect features but requires careful design to avoid overfitting. Designing and training ResNets for wafer defect classification involves balancing network depth and defect complexity. | ResNets have shown exceptional performance in image recognition, including high accuracy and precision in wafer defect classification, due to their deep learning capabilities without training difficulties. Their effectiveness relies heavily on the quality and diversity of the training data, highlighting the importance of a comprehensive dataset that encompasses all wafer defects. | ResNets architecture significantly boosts robustness, particularly in preventing overfitting, allowing for effective deep network training. This robustness is vital for tasks like wafer defect classification, where the model must adapt to unseen defects and pattern variations. ResNets also perform well with noisy data, common in semiconductor manufacturing, indicating high robustness. | Training and inference with ResNets demand significant computational and memory resources, posing challenges for real-time applications. Their performance hinges on the training dataset's quality and diversity, with inadequate representation of rare defects impairing defect detection accuracy. Furthermore, interpreting ResNet decisions is difficult, complicating the diagnosis of misclassifications and errors in defect categorization. |
| **Adversarial Training** | Wafer defect classification deals with a wide array of defect types. The complexity of adversarial training increases as it must generate adversarial examples that effectively mimic this diversity, challenging the model's ability to generalize across such a varied dataset. Adversarial training needs to navigate through this high-dimensional input space, complicating the generation and handling of adversarial examples. | Adversarial training can lead to models that are more robust to slight perturbations in input wafer images, potentially improving the detection accuracy for subtle, real-world wafer defects that could be missed by non-adversarial trained models. In a high-throughput environment like semiconductor manufacturing, this can be a bottleneck, necessitating a balance between model robustness and operational efficiency. | Adversarial training can fortify models against attempts to maliciously bypass or fool the defect detection system, enhancing the security of the manufacturing process. Manufacturing processes can introduce unanticipated variations in defect appearances. Adversarial trained models, by virtue of their exposure to a broader spectrum of input variations, may exhibit improved resilience to such natural variability. | ((1) The computational overhead required for generating adversarial examples and retraining models makes adversarial training resource-intensive. This can be particularly challenging in the context of wafer defect classification, where models may need to be frequently updated or retrained to adapt to new types of defects or changes in the manufacturing process, and (2) There's a risk that models become overly optimized for detecting adversarial examples at the expense of their ability to recognize genuine defects. |
| **XGBoost** | XGBoost has shown exceptional performance in various ML competitions and tasks, including wafer defect classification. Its ability to handle large and complex datasets with high dimensionality makes it well-suited for high-resolution wafer images. XGBoost can efficiently differentiate between normal and defective wafers and classify the types of defects with high accuracy, thanks to its robust handling of imbalanced data. | XGBoost demonstrates high performance in ML tasks such as wafer defect classification, effectively managing large, complex wafer maps with high dimensionality. Its proficiency in processing detailed, high-resolution images, crucial in semiconductor manufacturing, allows for accurate classification of wafer defects. This is attributed to its strong handling of imbalanced data and iterative improvement in areas of mistake. | The robustness of XGBoost in wafer defect classification comes from its gradient boosting mechanism, which focuses on correcting the errors of previously built models in the sequence of trees. It reduces bias and variance, leading to a more stable classification model. It includes built-in mechanisms for handling missing data to prevent overfitting, enhancing its robustness in applications where the quality of data can vary, such as in wafer defect detection. | (1) XGBoost requires a significant amount of labeled data to train effectively. In the context of wafer defect classification, obtaining a large and diverse dataset with accurately labeled defects can be challenging and costly, (2) Adaptability to New Wafer Defect Types: XGBoost models, once trained, may not readily adapt to new types of defects that were not present in the training data. This requires periodic retraining with updated datasets that include new defect types, which can be resource-intensive. |
| **Decision Tree** | Decision trees for wafer defect classification can vary in complexity based on the depth of the tree and the number of features considered. A deeper tree with more nodes can capture more detailed distinctions among different types of defects but may also lead to a more complex model that's harder to interpret. The complexity of decision trees can increase if the features extracted from wafer images or sensor data are high-dimensional. | Decision trees can achieve high accuracy in classifying wafer defects if the features selected for the nodes effectively differentiate between the types of defects. Decision trees are generally fast to train and predict, making them suitable for applications where rapid classification of wafer defects is required. However, the performance can be affected by the size of the dataset and the complexity of the tree structure. | Decision trees can handle non-linear relationships between features and classes well, which is beneficial in wafer defect classification where the relationship between sensor readings or image features and defect types might be complex. While decision trees are robust to outliers, they are prone to overfitting, especially if the tree is too deep. Overfitting can be mitigated by pruning the tree, setting a maximum depth, or using ensemble methods like Random Forests. | (1) decision trees are susceptible to overfitting when dealing with complex wafer defect classification problems. This can lead to poor generalization to unseen data, (2) Decision trees can be sensitive to small variations in the training data, leading to different tree structures. This sensitivity can affect consistency in classification performance, and (3) Their performance are highly depends on the choice of features. In wafer defect classification, if the features do not capture the nuances of different defect types, the tree may not perform well. |
| **Support Vector Machine (SVM)** | For wafer defect classification, the SVM's complexity is influenced by the dimensionality of the data (e.g., images' pixel intensity values) and the choice of the kernel. The training process can become computationally intensive for large datasets, common in semiconductor manufacturing due to high-resolution imaging techniques. The choice of kernel (linear, polynomial, RBF, etc.) plays a crucial role in managing the algorithm's complexity. | SVMs can achieve high accuracy in classifying wafer defects, especially when the data is not linearly separable. The use of appropriate kernel functions allows SVMs to efficiently handle complex defect patterns in semiconductor processes. The ability of SVMs to generalize well (i.e., for unseen data) is crucial for wafer defect classification. This stems from SVM's foundation on the principle of structural risk minimization, which aims to minimize an upper bound of the generalization error. | SVMs are relatively robust to noise in the data. Their reliance on support vectors (data points that are closest to the decision boundary) makes them less sensitive to outliers or non-defect-related variations in the wafer images. The regularization parameter in SVM helps prevent overfitting, a critical aspect when dealing with the high-dimensional data in wafer defect classification. This ensures that the model remains generalizable across different manufacturing batches or conditions. | (1) The performance of SVM heavily depends on the choice of the kernel and its parameters (e.g., C for regularization, gamma for the RBF kernel). Finding the optimal parameters can be challenging and time-consuming, requiring extensive cross-validation or grid search techniques, and (2) The computational cost of training SVM models can be high for large datasets, which is often the case in wafer defect classification due to the high volume of production and inspection processes. This can limit the real-time application of SVM in some scenarios |
| **K-Nearest Neighbor (KNN)** | (1) For wafer defect classification, the complexity of KNN largely depends on the dimensionality of the feature space. Wafer images may require high-dimensional feature vectors to capture defect characteristics, leading to computational complexity, and (2) The KNN algorithm itself is simple, with the primary operation being the computation of distances between feature vectors. In the context of wafer defect classification, the computational cost can become significant. | KNN can achieve high accuracy in wafer defect classification if the defects exhibit distinguishable patterns in the feature space and if an appropriate distance metric is chosen. The performance heavily relies on the selection of k, the number of neighbors. A well-chosen k can help the model effectively generalize from the training data to unseen data, but the optimal k might vary depending on the defect types and the distribution of the data. The presence of noise in wafer defect data can adversely affect KNN's performance. | KNN inherently handles non-linear data well, making it suitable for wafer defect classification where the relationship between features and defect types may not be linear. Its robustness to non-linearities allows it to capture complex defect patterns without the need for explicit model structuring, and (2) KNN's performance is influenced by the scale of features, as distance metrics are sensitive to feature magnitude. In wafer defect classification, features extracted from images may have varying scales, necessitating normalization | (1) As the size of the wafer dataset grows, KNN's computational cost becomes a significant limitation. Storing the dataset in memory and computing distances for each classification can be impractical for large wafer defect analysis, (2) Interpreting the model's decisions in the context of wafer defect classification can be challenging. It does not provide insight into the importance of different features in predicting defect types, making it difficult to derive actionable insights for process improvement, (3) The choice of distance metric significantly impacts KNN's effectiveness in classifying wafer defects |
| **Generative Adversarial Network** | (1) GANs consist of two main components: the generator and the discriminator. The complexity of the generator and the discriminator can vary based on the specific architecture (e.g., DCGAN, WGAN) and the depth and width of the neural networks employed. For wafer defect classification, the complexity is often higher due to the need for fine-grained feature detection, and (2) Training is computationally intensive and time-consuming, for high-resolution wafer images. | (1) GANs can achieve high accuracy in classifying wafer defects. The performance of GANs is heavily dependent on the quantity and quality of training data, (2) GANs can generate realistic synthetic images of wafer defects, which can be used to augment training datasets, especially when certain types of defects are rare or underrepresented. This can lead to improved model generalization and performance. | (1) GANs trained on diverse wafer datasets can generalize well to unseen wafer defect types or variations, making them robust to changes in defect patterns as manufacturing processes evolve, and (2) The robustness of GANs can be compromised if the training data is not representative of the actual distribution of wafer defects. Overfitting to the training data or failing to capture rare defect types reduces the effectiveness of the model in real-world applications. | (1) A common issue with GANs is mode collapse, where the generator starts producing a limited variety of outputs. In the context of wafer defect classification, this could mean failing to generate or recognize certain types of defects, impacting the diversity of the training dataset, and (2) GANs are notoriously difficult to train, with stability issues often arising due to the adversarial training dynamics. This can lead to prolonged training times and the need for meticulous tuning of hyperparameters. |
| **Random Decision Forests (RDF)** | (1) RDFs can handle the high dimensional data often found in wafer defect classification without feature selection. The complexity of the model grows with the number of trees and depth, allowing it to capture intricate patterns of defects on wafers, and (2) Training involves building multiple decision trees, which can be computationally intensive but are parallelizable. The complexity increases with the amount of data of high-resolution wafer images. | (1) RDFs are highly accurate for wafer defect classification. By aggregating the decisions of multiple trees, they reduce the risk of overfitting, a common issue with single decision trees, leading to better generalization on unseen data, (2) The ensemble nature of RDFs helps in handling the variance and biases of individual trees, making the model more generalizable to different types of wafer defects. | (1) RDFs are relatively robust to noise and outliers, which are common in wafer defect data due to variations in manufacturing and measurement errors. The ensemble approach helps mitigate the impact of noisy data on classification performance, and (2) RDFs can automatically assess the importance of different features for classification, which is crucial in wafer defect classification where not all features may be equally relevant to identifying specific defect types. | (1) In wafer defect classification, RDFs lack interpretability. Understanding why a particular decision or classification was made can be challenging, (2) RDFs can be memory-intensive, as they require storing numerous trees. This can be particularly challenging with the large datasets typical in wafer defect classification, and (3) Extremely high-dimensional spaces (such as those in high-resolution wafer images) may pose challenges for RDFs, requiring careful tuning and possibly dimensionality reduction techniques. |
| **CNN for Multi-Label Defect** | The model's complexity arises from: (1) To capture the intricate patterns of wafer defects, multiple convolutional layers are often required, increasing the model's depth and complexity, (2) The need for meticulous parameter tuning (e.g., filter sizes, strides, padding) to effectively learn wafer defect features without overfitting or underfitting, and (3) Preprocessing steps such as normalization, augmentation, and defect localization. | Multi-label CNNs offer high performance in wafer defect classification, attributed to their ability to learn spatial hierarchies of features directly from image data. Key performance aspects: (1) High classification accuracy for multiple defect types due to the network's capability to capture complex patterns and variations in defect appearances, (2) Good generalization to new, unseen wafer defect images, especially when trained on very large and diverse datasets. | The robustness of Multi-label CNNs in wafer defect classification is influenced by their capacity to handle variations and disturbances in the input data. They are: (1) Robust to variations in defect scale, orientation, and position due to the pooling layers and the hierarchical feature extraction process, (2) Effective in distinguishing defects from noise and other non-defective anomalies, even in low-contrast/poor-quality images, (3) The use of pre-trained models can enhance robustness by leveraging learned features. | (1) Training multi-label CNNs requires significant computational resources, including powerful GPUs and substantial memory, which may not be accessible to all organizations, (2) While CNNs excel at capturing spatial features, they may struggle with defects that are better characterized by non-visual properties (e.g., electrical or functional anomalies) unless integrated with other data sources or sensor inputs, and (3) The performance heavily relies on the availability of large and annotated datasets. Gathering such datasets is time-consuming and expensive |

# VI. EXPERIMENTAL EVALUATIONS

## A. Compiling Datasets for the Evaluations

Our study aimed to identify and classify common wafer defect patterns using a blend of real data from Samsung Electronics in Korea and synthetic data, based on the methodologies of DeNicolao et al. [92] and Jeong et al. [93]. This dual-data approach enhanced our evaluation of the algorithms. We focused on four main defect patterns: circle, cluster, repetitive, and spot, analyzing each using a probabilistic model to assess the likelihood of die failure at specific wafer locations. The real dataset from Samsung Electronics consists of 787 wafer maps collected from 45 lots. This included 11932 dice.

The simulated dataset is based on 400 wafer maps, with each map featuring 400 chips laid out in a 20x20 grid. These maps simulated five types of defect patterns: circle, cluster, repetition, spot, and Spatially Homogeneous Bernoulli Process (SHBP) [94]. Each is represented by 80 maps. We introduced varying noise levels in these maps, applying eight distinct levels from 0.05 to 0.4. For each combination of noise level and defect pattern, we generated ten wafer maps, resulting in datasets grouped by noise levels (0.05, 0.1, etc.). In total, these 400 wafer maps were segregated into four separate datasets. The simulation process was based on the method proposed by DeNicolao et al. [92]. Our spatial randomness test, using spatial lags of 3, showed that 98.8% of the SHBP wafer maps conformed to the anticipated patterns. A "spatial lag" is a measure that captures the influence of a defect characteristic (e.g., size, or type) at one location with its values at nearby locations. The "lag" denotes the distance between locations being compared. In contrast, all maps displaying spatial defect patterns were disqualified. This approach is in line with the methodologies of DeNicolao et al. [92] and Jeong et al. [93]. To realistically mimic common wafer defect patterns, we applied specific probabilistic models for each defect type, aimed at stimulating the failure locations on the wafer. The probabilistic formulas used to create the varied patterns of circle, cluster, repetitive, and spot defects are detailed in the equations in [9]. We present below equations that show the probabilistic expressions and the controlling parameters used to depict the position of a defective die on a simulated process zone:

- *Spot* - Let σ be the width, ($x_c$, $y_c$) be the coordinates of the wafer's center, and *r* be the distance between the defect centers and the wafer: $p(x,y) = \exp\left(r^2/2\sigma^2\right), \quad r^2 = (x-x_c)^2 + (y-y_c)^2$
- *Circle* - Let σ be the radius and ($x_c$, $y_c$) be the coordinates of its center: $p(x,y) = 1 - \exp\left(r^2/2\sigma^2\right), \; r^2 = (x-x_c)^2 + (y-y_c)^2$
- *Repetitive* - Let *T* and φ be the positioning of the row *T* and the column φ: $(\text{horizontal}): p(x,y) = \left(1+\sin\left(2\pi y/T + \phi\Phi\right)\right)/2$
  $$(\text{vertical}): p(x,y) = \left(1+\sin\left(2\pi x/T + \phi\Phi\right)\right)/2$$
- *Cluster* - Let "OR" and "AND" be logical operators:
  $$(\text{"}AND\text{"}): p(x,y) = p_1(x,y)y_2(x,y)$$
  $$(\text{"}OR\text{"}): p(x,y) = p_1(x,y) + p_2(x,y) - p_1(x,y)p_2(x,y)$$

We adopted the data imbalance rate as defined by He and Garcia [95], where the enhanced imbalance level is specified in equation 1. In this equation, $N_{maj}$ represents the total number of samples across all majority classes, while $N_{min}$ stands for the total number of samples across all minority classes. The data instances distribution are as follows:

- Non-defective (normal) wafer maps: 55% of the total dataset.
- Defective wafer maps distribution is as follows: Edge-ring: 20%, Scratch: 12%, Loc: 10%, Edge-loc: 9%, Center: 8%, Donut: 6%, Random: 5%, and Near-full: 2%.

$$r_{imb} = \sum N_{maj} / \sum N_{min} \tag{1}$$

In our experiments, we addressed missing data by substituting absent values with the average of the respective attribute. Furthermore, attributes that exhibited an excessive number of missing values were excluded from analysis. An additional crucial step in our data preprocessing involved the normalization of attribute values. Through z-score normalization [8], we adjusted these values so that the normalized attributes achieved mean values of 0 and standard deviations of 1, respectively.

## B. Evaluation Metrices

We utilized the subsequent metrics for assessment:

- *Classification accuracy (Acc):* It refers to the measure of the proportion of correct predictions made by a classification model. It is expressed as follows:

  Classification Accuracy = (Number of Correct Predictions) / (Total Number of Predictions)

- *Coefficient of determination ($R^2$):* It assesses the ability of a model to accurately elucidate and forecast future clustering outcomes. It is derived utilizing Equation 2.

  $$R^2 = 100 \times \left(1 - \frac{\sum_{i=1}^{n}(x_i - m_i)^2}{\sum_{i=1}^{n}(x_i - \bar{x})^2}\right) \tag{2}$$

  where $m_i$ is the predicted output

- *F1-measure:* It harmonizes precision and recall by generating a single score. The calculation of the F1-measure is achieved through the following equation:

  F1-measure= 2*(Precision * Recall) / (Precision + Recall) (3)

  Where, precision represents the ratio of true positive predictions to the total number of positive predictions, while recall denotes the ratio of true positive predictions to the total number of actual positive instances in data.

- *Adjusted Rand Index (ARI):* It quantifies the similarity between two data clustering, adjusting for the chance grouping of elements. The ARI adjusts for expected chance agreement.

  ARI = (RI - Expected_RI) / (Max_RI - Expected_RI) (4)

  RI (Rand Index) = (a + b) / (a + b + c + d), where *a* is the count of element pairs that are in the same subset in both clustering, *b* is the count of element pairs that are in different subsets in both clustering, *c* is the count of element pairs in the same subset in one clustering but in different subsets in the other, and *d* is the count of element pairs in different subsets in one clustering but in the same subset in the other. Expected_RI is the expected value of the Rand Index. Max_RI is the maximum possible value of the Rand Index.

- Clustering accuracy (γ): γ is computed by contrasting the projected cluster outcome with the real outcome.

  $$\gamma = \frac{\text{length}\left(X = \hat{x}\right)}{\text{length}\left(X\right)} \tag{5}$$

  $X$ is the correct value and $\hat{x}$ is the estimated one

- *Normalized Mutual Information (NMI):* It is a standardized measure ranging from 0 (no mutual information) to 1 (perfect match), used to evaluate the similarity of two clustering results, especially when the true classifications are known.

### C. Methodology for Selecting Representative Papers and Ranking the different Techniques and Sub-Categories

The following methodology was utilized in conducting the experimental evaluations:

- **Evaluating individual techniques:** After a comprehensive review of papers presenting algorithms employing specific techniques, we identified the paper with the greatest impact. The algorithm detailed in this influential paper was chosen as the representative for its respective technique. To determine the most significant paper among those reporting algorithms using the same technique, we considered various factors: level of innovative contribution, advancement in the state of the art, publication date/recentness, and number of citations.
- **Ranking the techniques**: We calculated the mean scores of the selected algorithms which made use of the same technique. Then, we ranked these techniques that are part of the same main sub-category, according to their scores.
- **Ranking the sub-categories:** We calculated the mean scores of the chosen algorithms that operated under a common sub-category. Subsequently, these sub-categories were ranked according to their scores.

### D. Evaluation Setup

#### *1) Single-Label algorithms:*

- Common Parameters:
  - Learning Rate: 0.001; Batch Size: 32; Epochs: 100; Optimizer: Adam.
- Unique Parameters by Model:
  - *Residual Neural Network (ResNet)*: Layers: ResNet-50.
  - *CNN*: Number of Convolutional Layers: 3 layers with increasing filter sizes 64, 128; Kernel Size: 3x3.
  - *Adversarial Training*: Epsilon: 0.01; Perturbation Iterations: 3 iterations for generating examples.
  - *XGBoost*: max_depth: 6; n_estimators: 100; subsample: 0.8.
  - *Adaptive Boosting*: n_estimators: 50; Learning Rate: 1.0.
  - *SVM*: Kernel: 'rbf'; C: 1.0; Gamma: 'scale'.
  - *KNN*: n_neighbors: 5; Metric: 'minkowski'; p: 2 (Euclidean distance).
  - *LVQ*: Number of codebooks: 10% of the training set size.
  - *Hopfield Network*: Network size: Equivalent to the feature size of the input data; Update rule: Asynchronous.

#### *2) Multi-Label algorithms*

- Common Parameters:
  - Regularization: L2; Learning Rate: 0.001; Batch Size: 32; Epochs: 100.
- Unique Parameters by Model:
  - *GANs*: ***For the generator***: Latent dimension size: 100; number of layers: 5; Type of layers: Deconvolution layers; Activation Function: LeakyReLU activation for intermediate layers, and a Tanh or Sigmoid activation function for the output layer. ***For the discriminator:*** Number of layers: 5; Activation Function: LeakyReLU.
  - *RDFs*: 100 trees with no maximum depth.

#### *3) Single-Agent algorithms*

- Common Parameters:
  - Number of Clusters (for HAC and K-Means) / States (for HMT): 5.
  - Distance Metric (for HAC and K-Means): Euclidean distance.
  - Initialization Method (for K-Means): K-means++.
  - Convergence Criteria: maximum number of iterations 300.
- Unique Parameters by Model:
  - *HMT*: Convergence Criteria: Improvement in log-likelihood.
  - *HAC*: Linkage Criteria: Average Linkage.

#### *4) Multi-Agent algorithms*

- Common Parameters:
  - Bandwidth (for Mean Shift): starting with 0.5.
  - Epsilon and MinPts (for Density-Based Methods): starting point for Epsilon is the average distance between points, and MinPts starting at 2 *.
  - Initialization (for Mean Shift): k-means++.

#### *5) Single-Type algorithms*

- Common Parameters:
  - Batch Size: 32; Epochs: 100; Optimizer: Adam.
- Unique Parameters by Model**:**
  - *CNN:* Architecture: 3 convolutional layers, each followed by max-pooling layers; Filters: Start with 32 filters in the first layer and double the number in each subsequent layer; Kernel Size: Use 3x3 kernel size for all convolutional layers.
  - *GAN*: Noise vector size: 100, for the noise vector input to the generator.

#### *6) Multi-Type algorithms*

- Common Parameters:
  - Learning Rate: 0.001; Batch Size: 32; Epochs: 100.
- Unique Parameters by Model:
  - CNN Architecture: AlexNet.
  - DNN Architecture: Each layer has a ReLU activation function, with the final layer using a softmax activation for classification.

### E. The Experimental Results

For the representative papers, we developed our own implementations using TensorFlow, as described by Sinaga and Yang [96]. We trained these implementations using the Adam optimizer, as suggested by Sinaga and Yang [96]. TensorFlow's APIs provide users with the flexibility to create their own algorithms [97]. Our development language was Python 3.6, and we utilized TensorFlow 2.10.0 as the backend for the models. Fig. 11 shows the results for the label-based, agent-based, and type-based techniques, respectively. These tables also include the rankings of the techniques within the same sub-category and the rankings the sub-categories.

We applied a One-way ANOVA parametric test [98] to examine whether the discrepancies in individual accuracy scores for each technique are significant enough to be considered statistically meaningful, and to assess if the variations in accuracy among different techniques reach statistical significance. The outcomes are detailed in Table 2.

| Sub Category | Technique | Selected Papers | Metric | Score % | Technique Rank | Sub Category Rank |
|---|---|---|---|---|---|---|
| Single-Label | Residual Neural Network | [24] | Acc | 86.42 | 1 | 1 |
| | | | $R^2$ | 89.78 | | |
| | | | F1 | 85.83 | | |
| | Convolutional Neural Network | [19] | Acc | 84.97 | 2 | |
| | | | $R^2$ | 90.04 | | |
| | | | F1 | 83.98 | | |
| | Adversarial Training | [27] | Acc | 82.48 | 3 | |
| | | | $R^2$ | 84.96 | | |
| | | | F1 | 79.68 | | |
| | Networks with Self-Calibrated Convolutions | [48] | Acc | 82.17 | 4 | |
| | | | $R^2$ | 83.52 | | |
| | | | F1 | 78.80 | | |
| | XGBoost | [29] | Acc | 81.46 | 5 | |
| | | | $R^2$ | 82.46 | | |
| | | | F1 | 79.97 | | |
| | Adaptive Boosting | [53] | Acc | 80.03 | 6 | |
| | | | $R^2$ | 81.12 | | |
| | | | F1 | 78.04 | | |
| | Support Vector Machine | [36] | Acc | 80.93 | 7 | |
| | | | $R^2$ | 81.54 | | |
| | | | F1 | 77.04 | | |
| | K-Nearest Neighbors | [44] | Acc | 77.69 | 8 | |
| | | | $R^2$ | 79.63 | | |
| | | | F1 | 76.65 | | |
| | Learning Vector Quantization | [46] | Acc | 74.52 | 9 | |
| | | | $R^2$ | 75.06 | | |
| | | | F1 | 71.37 | | |
| | Hopfield Artificial Neural Network | [50] | Acc | 71.83 | 10 | |
| | | | $R^2$ | 74.48 | | |
| | | | F1 | 68.33 | | |
| Multi-Label | Generative Adversarial Network | [55] | Acc | 84.16 | 1 | 2 |
| | | | $R^2$ | 88.85 | | |
| | | | F1 | 83.77 | | |
| | Support Vector Machine | [59] | Acc | 80.56 | 2 | |
| | | | $R^2$ | 81.37 | | |
| | | | F1 | 79.51 | | |

(a)

| Sub Category | Technique | Selected Papers | Metric | Score % | Technique Rank | Sub Category Rank |
|---|---|---|---|---|---|---|
| Agent-Label | Hidden Markov Tree | [62] | ARI | 76.1 | 1 | 2 |
| | | | γ | 83.9 | | |
| | | | NMI | 75.2 | | |
| | Density-Based Methods | [66] | ARI | 69.5 | 2 | |
| | | | γ | 79.4 | | |
| | | | F1 | 67.9 | | |
| | Mean Shift | [64] | ARI | 63.9 | 3 | |
| | | | γ | 73.8 | | |
| | | | NMI | 62.6 | | |
| Multi-Agent | Statistical Hierarchical Based Methods | [73] | ARI | 78.6 | 1 | |
| | | | γ | 89.4 | | |
| | | | NMI | 76.7 | | |
| | Hierarchical Agglomerative Clustering | [71] | ARI | 73.2 | 2 | 1 |
| | | | γ | 81.6 | | |
| | | | NMI | 71.6 | | |
| | K-Means-Based Methods | [75] | ARI | 68.4 | 3 | |
| | | | γ | 70.9 | | |
| | | | NMI | 67.7 | | |

(b)

| Sub Category | Technique | Selected Papers | Metric | Score % | Technique Rank | Sub Category Rank |
|---|---|---|---|---|---|---|
| Single-Type | Convolutional Neural Network | [77] | Acc | 88.47 | 1 | 1 |
| | | | $R^2$ | 92.63 | | |
| | | | F1 | 87.04 | | |
| | Convolutional Generative Adversarial Network | [81] | Acc | 87.92 | 2 | |
| | | | $R^2$ | 90.58 | | |
| | | | F1 | 85.74 | | |
| | K-Means | [82] | Acc | 80.83 | 3 | |
| | | | $R^2$ | 85.06 | | |
| | | | F1 | 78.49 | | |
| Multi-Type | CNN for Multi-Type | [88] | Acc | 83.25 | 1 | 2 |
| | | | $R^2$ | 86.78 | | |
| | | | F1 | 81.58 | | |
| | Deep Neural Network | [91] | Acc | 84.86 | 2 | |
| | | | $R^2$ | 88.69 | | |
| | | | F1 | 83.07 | | |

(c)

**Fig. 11:** The scores of (a) label-based, (b) agent-based, and (c) type-based wafer defect classification techniques. Also, the table show the ranking of the techniques within their respective sub-categories and the ranking of the sub-categories.

**TABLE 2:** ONE WAY ANOVA STATISITCAL TEST FOR THE DISCPENCAY OF THE ACCURACY SCORES WITHIN AND AMONG THE TECNIQUES

| | Within Techniques | | Between Techniques | | | |
|---|---|---|---|---|---|---|
| Techniques | Sum of Square (SS) | Mean Square (MS) | Sum of Square (SS) | Mean Square (MS) | F-Statistic | p-value |
| Single-Label | 78.2 | 18.16 | 247.4 | 42.8 | 1.71 | 0.0813 |
| Multi-Label | 2.633 | 0.1557 | 32.436 | 8.109 | 3.08 | 0.0462 |
| Single-Agent | 0.5375 | 2.6404 | 372.808 | 16.404 | 6.21 | 0.0025 |
| Multi-Agent | 0.4854 | 13.02 | 260.4 | 124.2 | 5.2400 | 0.00073 |
| Single-Type | 1.3055 | 0.1088 | 181.5124 | 90.7562 | 834.16 | 0.00006 |
| Multi-Type | 0.769 | 0.0961 | 6.4803 | 6.4803 | 67.4329 | 0.00021 |

### F. Discussion of the Experimental Results

1. Convolutional Neural Network

The method demonstrated superior precision in identifying and categorizing wafer map defects, particularly when handling extensive datasets. It was notably proficient at interpreting the geographical distribution of flaws on the wafer surface, a key element in detecting defective patterns. Compared to this method, traditional machine learning approaches fell short when faced with intricate patterns or when there was a wealth of labeled training data available. Its inherent translational invariance enabled it to recognize patterns irrespective of their location within the image. This characteristic is advantageous in detecting defects on wafer maps, as flaws could appear anywhere on the wafer. The method's hierarchical learning ability, which allows it to grasp low-level attributes in initial layers and high-level attributes in later layers, strengthened its competency in discerning complex patterns.

2. Residual Neural Network-Based Classification

ResNets, with its deep learning capabilities, excelled in identifying complex patterns in wafer maps, outperforming traditional machine learning and standard CNNs. However, it sacrificed interpretability and computational efficiency. Its ability to counter vanishing gradients led to consistent loss reduction, indicating efficient learning. But the training demands surpassed simpler methods due to ResNets' complexity, with training duration and resources being influenced by ResNet depth and data volume. The duration required for training were directly affected by the depth of the ResNet and the volume of wafer map data. The impact of weight initialization and hyperparameter selection was significantly noticeable in the performance of ResNets.

3. Generative Adversarial/Adversarial Training Classification

Adversarial training improved model robustness by making them less sensitive to input variations and noise. It also enhanced the model's ability to generalize from training data to real-world situations. However, this method was computationally demanding, requiring more resources and time. In comparison, GANs focused on creating new data for training, while adversarial training strengthened the model's resilience to variations and perturbations. Combining these methods can be beneficial, as GANs can expand the dataset and adversarial training can ensure robustness against the new data's variations. In industries like semiconductor manufacturing, these methods can reduce costs in quality control by automating defect classification, and they can be customized to target specific types of defects for increased efficiency.

3. XGBoost-Based Classification

XGBoost excelled in performance and flexibility, although careful tuning and preprocessing were required, given the complexity of wafer map data. The model's predictions were somewhat challenging to interpret. XGBoost outperformed other gradient boosting methods with a quicker path to minimum error, faster convergence, and optimized computations for increased speed and lower computational costs. It efficiently handled missing data, significantly reducing preprocessing time. The algorithm detected and learned from non-linear patterns and prevented overfitting using various regularization penalties.

4. Decision Tree-Based Classification

The algorithms rooted in classical methods, specifically those utilizing gradient-boosted decision trees, demonstrated high effectiveness in correctly detecting flawed patterns in wafer maps and dealing with missing data. This resulted in commendable accuracy rates. The ability of the gradient boosting framework to efficiently optimize complex loss functions is seen as the reason for this prediction advantage. This framework also has a built-in mechanism to deal with absent values and a structured method for comprehending the significance of various features in predictions. Experimental results showed that these algorithms tend to overfit on smaller datasets, with their improved performance linked to prioritizing instances with large gradients.

5. Random Decision Forests-Based Classification

Generally, the RDF technique accurately recognized most wafer maps in the test group as defective or non-defective. Yet, its accuracy faltered with imbalanced datasets where one class significantly outstripped the other. Experiment outcomes showed the RDF method's high efficiency, even with minor hyperparameter tweaks. It stood out in dealing with skewed datasets, commonly seen in defect detection where non-defective wafers far exceed defective ones. The approach adeptly managed non-linear feature interactions and detected feature interplays, with its ensemble nature making it more resistant to overfitting.

6. Support Vector Machine

Compared to Decision Trees, Neural Networks, and Random Forests, the SVM technique showed higher accuracy and generalization in some cases for identifying faulty wafer map patterns due to its proficiency in handling high-dimensional data. But it struggled with high complexity, difficulty processing extremely large datasets, lack of interpretability, and needed careful parameter calibration. Its strength was in its flexibility in handling diverse data patterns, which was achieved by using various kernel functions for creating non-linear decision boundaries and complex data transformations. SVM's kernel trick models non-linear boundaries for complex defect detection, and its regularization parameter prevented overfitting.

7. K-Nearest Neighbor-Based Classification

The technique achieved decent precision rates by integrating distance-related classification and normalizing the dataset used for training. However, the presence of unrelated features and inconsistent feature scaling significantly hindered the method's efficiency. The model's detection capabilities varied across distinct types of defects, excelling at identifying specific kinds due to the distinct distribution and density of various defect types within the feature space. The method was also computationally demanding when processing large datasets, as it required the calculation of the distance between a given test point and all points in the dataset for prediction purposes. Although the technique could predict the class label, it didn't offer any measure of confidence for that prediction.

8. Learning Vector Quantization (LVQ) Classification

The LVQ method outperformed clustering in terms of accuracy due to its ability to harness label information. Nonetheless, it necessitated extensive data preparation, given that labels are a prerequisite for training data. Deep learning techniques such as CNNs surpassed the LVQ method in terms of accuracy by identifying more intricate patterns within the data. However, these methods were more challenging to decipher, required greater data volumes, and were computationally demanding. LVQ effectively handled noisy wafer map data and exceled in complex, non-linear classification tasks. Its learning rate and parameters can be optimized, but it required a large amount of labeled training data.

## VII. POTENTIAL FUTURE PERSPECTIVES

We present in this section some potential future improvements for identifying the defective patterns in WBM using classification machine learning techniques.

### A. Deep Learning-Based Classification

#### 1. Artificial Neural Network (ANN)-Based Techniques

- Synthetic Wafer Map Data Generation: Implement GANs to create artificial wafer map data. This approach is vital in wafer defect classification to enhance the diversity and volume of training datasets, thereby improving the robustness of defect detection models.
- Hyperparameter Tuning: Employ sophisticated methods like Bayesian optimization and evolutionary algorithms for hyperparameter tuning of neural networks. In wafer defect classification, this automated tuning is essential to optimize the performance of models, ensuring they are finely adjusted to the nuances of wafer defect patterns.
- Reinforcement Learning: Implement reinforcement learning algorithms to enable continuous performance enhancement of wafer defect classification models. This adaptive approach allows the models to dynamically adjust to changes and new patterns in the semiconductor manufacturing process, ensuring accuracy and effectiveness in defect identification.

#### 2. Convolutional Neural Network (CNN) Classifications:

- Advanced Convolutional Layers: Incorporate structures like dilated and depth wise separable convolutions in CNNs. These advanced convolutional techniques are effective for capturing the nuanced details in wafer images, optimizing the network's learning ability for intricate defect patterns.
- Ensemble Learning: Utilize ensemble learning by aggregating predictions from various CNN architectures. This approach improves the overall accuracy in wafer defect identification, as different models may capture different aspects of the semiconductor wafer defect data.
- Innovative Training Techniques: Employ advanced training methods like cyclic learning rates and knowledge distillation. These strategies enhance the efficiency of CNNs in learning from semiconductor defect data, particularly useful in scenarios with a large complex data.
- Automated Optimization: Apply AutoML tools to automate the selection of the most suitable CNN architecture and hyperparameters.
- Transfer Learning: Pretrain CNN models on large, diverse datasets and fine-tune them on specific wafer defect data. This is particularly effective in scenarios where defect data is limited, allowing CNN to learn general features first and then adapt to the specific characteristics of wafer defects.
- Customized CNN Architectures: Tailor CNN architectures by integrating attention modules and diverse layer configurations. These customizations enable focused, accurate identification of wafer defects, ensuring more precise and reliable classification.

#### 3. Residual Neural Network (ResNet) Classifications

- Enhanced Network Architecture: Integrate depth-wise separable convolutions into ResNet, optimizing it for the intricate and varied patterns found in wafer defects. This aims to improve the network's ability to learn from complex wafer defect data.
- Transfer Learning: Adapt ResNet with pre-training on datasets and tasks that are closely related to wafer defects. This enhances the model's ability to recognize and classify unique defect patterns effectively.
- Few-shot Learning: Implement few-shot learning techniques in ResNet to address scenarios with scarce examples of certain defect types. This is crucial for wafer manufacturing, where some defects are rare.
- Multi-scale Feature Extraction: Incorporate multi-scale feature extraction in the ResNet architecture. This allows the network to detect wafer defects of different sizes and shapes, a requirement in wafer defect classification.
- Unsupervised Learning: Use autoencoders for unsupervised learning to recognize standard wafer patterns and anomalies. This helps the network distinguish irregularities, which are classified using a trained ResNet, improving defect detection.
- Attention Mechanisms: Incorporate Transformer-like attention mechanisms into ResNet to focus on key areas of wafer maps. This enhances defect identification and classification in large, complex wafers.

#### 4. Generative Adversarial Network (GAN) Utilization

- Customized GAN Architectures: Innovations in GAN architectures, such as StyleGANs and BigGANs, offer significant potential for improving wafer defect detection. Tailoring these GANs to the specific nuances of wafer defect detection could lead to a considerable enhancement in performance. This involves adapting the network architecture and training process to suit wafer defects.
- Conditional GAN Application: The use of Conditional GANs, which incorporate additional information such as the type or location of defects into both the generator and discriminator, can refine wafer defect detection accuracy.
- Diverse Data Augmentation: Utilize CycleGANs in training for wafer defect detection to address rare or underrepresented defects. This increases defect type variety in training, broadening the model's learning of defect patterns and boosting detection effectiveness.
- Hybrid Models: Merge GANs with techniques like reinforcement learning or attention mechanisms for improved wafer defect detection.
- Few-shot Learning: Enabling learning from limited wafer defect examples or category descriptions aids in detecting rare defects, allowing for efficient training.
- Emphasis on Multi-scale and Hierarchical Feature Analysis: Future advancements in GANs could emphasize using multi-scale and hierarchical features to improve wafer defect identification accuracy. This method entails examining defects at different scales and detail levels.

#### 5. Adversarial Training-Based Classification

- Advanced Adversarial Strategies: Developing sophisticated adversarial attack methodologies, such as Fast Gradient Sign Method (FGSM) and Projected Gradient Descent (PGD), is crucial for cultivating highly resilient models in wafer defect detection. These methods should be tuned to generate large adversarial examples.
- Adversarial Defense Techniques Tailored for Semiconductor Applications: It is imperative to refine and

customize adversarial defense techniques, including adversarial training, defensive distillation, and feature squeezing, specifically for the domain of wafer defect detection. This involves modifying techniques to better identify and counteract simulated wafer defect attacks, enhancing the model's defense against deceptive inputs.

- *Multi-modal Training data*: Enhancing adversarial training to include multi-modal data, like optical and electron microscope images, improves the robustness of wafer defect classification models against data anomalies.
- *Incorporating Uncertainty Quantification*: Implementing uncertainty quantification in adversarial training is crucial for improving anomaly detection in wafer manufacturing. It quantifies model prediction confidence against adversarial examples, ensuring reliable assessment.
- *Robust Optimization*: Robust optimization strategies can improve a model's ability to generalize from adversarial examples to unseen data, which is crucial in the dynamic wafer manufacturing where new defect types may arise.
- *Application of GANs*: Utilizing GANs in adversarial training enhances model performance, with the generator creating complex simulated wafer defects and the discriminator focusing on their detection, thus boosting the model's accuracy in identifying real wafer defects.

**B. Traditional-Based Classification**

**1. XGBoost-Based Classification**

- *Hyperparameter Optimization*: For wafer data, implements Bayesian optimization and AutoML to fine-tune XGBoost parameters like learning rate, tree depth, and estimators.
- *Multimodal Learning Approach*: Integrates diverse data types (image and time-series sensor data) in a unified framework to enhance defect classification.
- *Ensemble Strategies*: Combines XGBoost with other models using stacking, bagging, and boosting to capture a wide range of wafer defect types.

**2. Decision Tree (DT)-Based Classification**

- *Integration of DL Techniques*: Merges CNNs and RNNs with decision trees to improve wafer defect classification.
- *Enhancement of IoT*: Integrates decision trees with IoT for real-time learning and defect prediction.
- *Ensemble Methods*: Uses Random Forest, Gradient Boosting, and AdaBoost for more accurate models.
- *Decision Tree Fusion*: Enhances high-dimensional data handling and feature extraction for classification.
- *Advancements in DT Algorithms*: Optimizes splitting criteria, pruning techniques, and data management.
- *Development of Multi-objective Decision Trees*: Focuses on accuracy, depth, and interpretability.

**3. Random Decision Forests (RDF)-Based Classification**

- *Hyperparameter Optimization*: Employing sophisticated hyperparameter optimization techniques, such as Bayesian Optimization or Genetic Algorithms, is essential in wafer defect identification. This refines critical hyperparameters, including tree count, maximum depth, and feature divisions, in RDFs. Such optimization ensures the model is finely tuned to the characteristics of wafer defect data.
- *Integration of Time-Series Data:* RDFs, when adapted to consider time-based correlations within the data points of wafers, can enhance their predictive performance. This acknowledges the dynamic nature of wafer manufacturing and the evolution of defect patterns over time.
- *Model Hybridization:* The amalgamation of RDFs with other advanced ML or deep learning models can create a more robust collective model for wafer defect classification. This hybridization strategy leverages the strengths of various algorithms, leading to heightened prediction accuracy.

**4. Support Vector Machine (SVM)-Based Classification**

- *Optimization of Kernel Functions:* Tailor kernel functions for wafer maps to enhance SVM performance in defect detection. Focus on creating or improving kernels for better classification accuracy.
- *Hyperparameter Tuning:* Utilize advanced methods like grid search, random search, or Bayesian optimization to fine-tune SVM hyperparameters. Aim to boost defect classification accuracy and efficiency in wafers.
- *Integration of SVM:* Merge SVMs with deep learning, especially CNNs, for enhanced feature extraction. This is adept at identifying complex wafer defects.
- *Ensemble Techniques:* Build high-precision models by combining multiple SVMs using ensemble methods like bagging or boosting. Each SVM focuses on different data subsets or features, improving classification robustness.

**5. Logistic Regression (LR)-Based Classification**

- *Hybrid Modeling:* Combining LR with advanced machine learning techniques like decision trees or neural networks improves defect classification in wafers. This blend leverages LR's simplicity and interpretability with the complex pattern recognition of advanced models, offering a powerful solution for detecting intricate wafer defects.
- *Enhanced Regularization Techniques:* Advances in L1 and L2 regularization significantly refine LR models for wafer defect classification. These methods address overfitting, boosting the model’s generalization, and ensuring high precision in detecting various defect types.
- *Integrating IoT:* Leveraging IoT technology for real-time data collection and analysis in wafer manufacturing enhances logistic regression models. This allows for immediate defect detection and classification, improving manufacturing efficiency and reliability. Continuous updates from IoT data keep the model current in identifying defect patterns.

**6. K-Nearest Neighbor (KNN)-Based Classification**

- *Weighted KNN:* Utilize a weighted KNN algorithm to predict defects in wafers. Weight neighbors based on their distance, giving more influence on closer neighbors. This increases defect classification precision by considering the spatial relevance of each neighbor.
- *Adaptive KNN for Variable Data Density:* Develop an adaptive KNN algorithm that varies its k-value according to the data density in wafer datasets. This flexibility is key to improving accuracy, especially in areas with sparse data, where fixed-k methods may falter.
- *Unsupervised Learning:* Employ unsupervised learning alongside KNN to identify new defect patterns in wafers. This technique is useful for spotting unusual or emerging defects, enhancing the defect detection process's comprehensiveness and robustness.

## VIII. CONCLUSION

ML algorithms have proven highly capable in wafer defect detection, despite the lack of a comprehensive review in this field. In this survey paper, we amalgamate existing studies to highlight the strengths, limitations, and potential applications of different ML classification algorithms in defect detection on wafer maps. We reviewed algorithms utilizing distinct techniques, sub-categories, and categories, providing a classification system to facilitate algorithm comparison and to guide future research.

This survey not only presented a detailed framework for categorizing wafer defects algorithms but also included *Observational* and *experimental* evaluations to measure the effectiveness of different approaches. Our *Observational evaluation* focused on ML classification techniques for identifying defect patterns in wafer maps based on five criteria. Through *experimental evaluation*, we compared and ranked various methodology categories and techniques, including those utilizing the same technique, different techniques within the same sub-category, different sub-categories within the same category, and different categories. Based on our experimental results, the CNN-Based classification was superior, especially with large datasets. It excelled at interpreting wafer surface imperfections and recognizing patterns regardless of their image location due to its hierarchical learning ability and translational invariance.

Below, we summarize the key findings from our experimental results:

- *Advantages of Deep Learning:* CNNs and ResNets demonstrate superior defect detection in wafer maps, outperforming traditional methods with their ability to analyze complex patterns and defect distributions. Their success is attributed to features like translational invariance and hierarchical learning.
- *Computational Efficiency and Interpretability:* Despite their high performance, deep learning models require significant computational resources and sacrifice interpretability, highlighting the need for efficient management in real-world applications.
- *Robustness through Adversarial and Generative Training:* GANs and adversarial training methods boost model robustness and generalization, essential for automating defect classification and reducing quality control costs, albeit with increased computational demands.
- *Performance of Traditional Machine Learning Models:* Traditional models like XGBoost, decision trees, RDFs, and SVMs have shown effectiveness in defect detection, each offering unique advantages in handling data complexity, despite some requiring extensive resources or presenting interpretability challenges.
- *Limitations of KNN and LVQ:* KNN and LVQ provide decent precision but lag behind deep learning in accuracy and complexity handling, with KNN affected by feature relevance and LVQ demanding significant data preparation.

Below, we present key insights and relevant conditions that illustrate the trade-offs between CNN-based, ML-based, and statistical approaches:

- ***CNN-based Approaches:***
  - Pros*:*
    1) *High Precision*: Especially effective with large datasets and intricate defect patterns, leveraging spatial relationships on the wafer surface.
    2) *Complex Pattern Recognition:* Ability to learn hierarchical features, from low-level to high-level attributes, enhancing defect detection capability.
    3) *Translational Invariance:* Can identify defects regardless of their position on the wafer, crucial for wafer map analyses.
  - Cons:
    1) *Interpretability:* These models are generally less interpretable than ML-based or statistical approaches, making it difficult to understand decision-making processes.
    2) *Computational Demand:* High computational resources and time are required, especially for training deep models like ResNets.
    3) *Generalization Issues:* While highly effective on the datasets they're trained on, CNNs might not generalize well to new, unseen wafer patterns without retraining.
- ***ML-based Approaches (XGBoost, Decision Trees, Random Decision Forests, SVM, K-Nearest Neighbors, and LVQ)***
  - Pros:
    1) *Flexibility and Performance:* Techniques like XGBoost and SVM offer robust performance and can handle non-linear patterns effectively.
    2) *Handling of Missing Data:* Certain ML approaches efficiently manage missing data, reducing preprocessing time.
    3) *Feature Importance Understanding:* Models like decision trees provide insights into the significance of different features in predictions.
    4) *Versatility in Application:* The variety of ML approaches allows for tailored solutions that can address specific types of defects or data characteristics, offering a bespoke approach to defect detection.
  - Cons:
    1) *Data Preparation and Tuning:* Require careful tuning and preprocessing, particularly for complex wafer map data.
    2) *Overfitting on Datasets:* Some ML methods may overfit when trained on limited data.
    3) *Computational Efficiency:* While generally less demanding than CNN-based methods, large datasets suffer computational challenges.
- **Statistical Approaches**
  - Pros:
    1) *Interpretability:* Higher than deep learning models, allowing for easier understanding of how decisions are made.
    2) *Lower Computational Demands:* Typically require less computational power and resources than complex ML or CNN-based models.
    3) *Effectiveness with Small Datasets:* Can perform well in scenarios with limited data, without the risk of overfitting inherent to more complex models.
  - Cons:
    1) *Limited in Handling Complex Patterns:* May not match the precision of CNNs or advanced ML techniques in detecting intricate defect patterns.
    2) *Lack of Spatial Recognition:* Unlike CNNs, statistical methods might struggle with spatial relationships and translational invariance.

## REFENCES

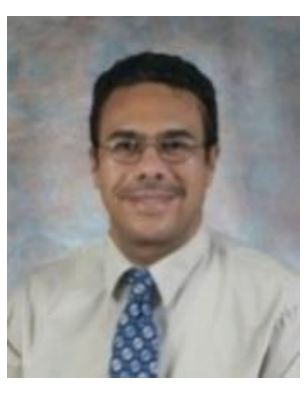

**Kamal Taha** has been an Associate Professor in the Department of Electrical Engineering and Computer Science at Khalifa University, UAE, since 2010. He received his Ph.D. in Computer Science from the University of Texas at Arlington, USA. He has over 100 refereed publications that have appeared in prestigious top ranked journals, conference proceedings, and book chapters. Over 30 of his publications have appeared in IEEE Transactions journals. He was an Instructor of Computer Science at the University of Texas at Arlington, USA, from August 2008 to August 2010. He worked as Engineering Specialist for Seagate Technology, USA, from 1996 to 2005 *(Seagate is a leading computer disc drive manufacturer in the US)*. His research interests span defect characterization of wafers, information retrieval, data mining, information forensics & security, bioinformatics, and databases, with an emphasis on making data retrieval and exploration in emerging applications more effective, efficient, and robust. He serves as a member of the editorial board for many journals and conferences.